\documentclass[letterpaper, 10 pt, conference]{ieeeconf}  

\IEEEoverridecommandlockouts                              

\usepackage[
top    = 54pt,
bottom = 58pt,
left   = 54pt,
right  = 54pt]{geometry}
\usepackage{booktabs}
\usepackage[font=small]{caption}
\usepackage{subcaption}
\usepackage{amsmath}
\usepackage{graphicx}
\usepackage{amssymb}
\usepackage{wrapfig}
\usepackage{chngpage}
\usepackage{etoc}
\usepackage[dvipsnames,rgb]{xcolor}

\usepackage[normalem]{ulem} 
\newcommand\hl{\bgroup\markoverwith{\textcolor{yellow}{\rule[-.5ex]{2pt}{2.5ex}}}\ULon}
\renewcommand\hl[1]{#1} 
\newcommand{\mathcolorbox}[2]{\colorbox{#1}{$\displaystyle #2$}}
\renewcommand{\mathcolorbox}[2]{#2} 

\usepackage{tikz}
\usepackage{multirow}
\usepackage{algorithm}
\usepackage{algpseudocode}
\usepackage{balance}
\usepackage{bm}
\usepackage{multirow}
\usepackage{siunitx} 
\usepackage{breakurl}
\usepackage{url}

\newcommand{\sd}[1]{} 

\newcommand{\beginsupplement}{%
    \setcounter{table}{0}
    \renewcommand{\thetable}{S\arabic{table}}%
    \setcounter{figure}{0}
    \renewcommand{\thefigure}{S\arabic{figure}}%
 }

\newcommand{\fig}[1]{Fig.~\ref{#1}}

\newcommand{\sect}[1]{Sec.~\ref{#1}} 
\newcommand{\etal}[0]{{\em et al.~}}

\definecolor{linkcolor}{RGB}{6,69,173}
\usepackage[
colorlinks=true,allcolors=linkcolor,pageanchor=true,
plainpages=false,pdfpagelabels,bookmarks,bookmarksnumbered,
backref=page
]{hyperref}

\usepackage{xspace}
\newcommand{\ngdf}{NGDF\xspace}
\newcommand{\ngdfs}{NGDFs\xspace}
\newcommand{\omgclr}{blue}
\newcommand{\cbrrtclr}{ForestGreen}
\newcommand{\ngdfclr}{orange}
\newcommand{\bclr}{purple}

\newcommand{\figw}{0.115}

\newcommand{\appfig}[1]{Appendix Fig.~\ref{#1}}
\newcommand{\app}[1]{App.~\ref{#1}} 

\newcommand{\appref}[1]{#1}

\newcommand{\arxivref}[1]{#1}

\title{\LARGE \bf 
Neural Grasp Distance Fields for Robot Manipulation
}

\author{Thomas Weng$^{1,2}$, David Held$^{2}$, Franziska Meier$^{1}$, and Mustafa Mukadam$^{1}$\\[2mm]
$^{1}$Meta AI, $^{2}$Carnegie Mellon University
}

\begin{document}

\maketitle
\thispagestyle{empty}
\pagestyle{empty}

\begin{abstract}
We formulate grasp learning as a neural field and present Neural Grasp Distance Fields (\ngdf). Here, the input is a 6D pose of a robot end effector and output is a distance to a continuous manifold of valid grasps for an object. 
In contrast to current approaches that predict a set of discrete candidate grasps, the distance-based \ngdf representation is easily interpreted as a cost, and minimizing this cost produces a successful grasp pose.
This grasp distance cost can be incorporated directly into a trajectory optimizer for joint optimization with other costs such as trajectory smoothness and collision avoidance. 
During optimization, as the various costs are balanced and minimized, the grasp target is allowed to smoothly vary, as the learned grasp field is continuous. 
\hl{We evaluate \ngdf on joint grasp and motion planning in simulation and the real world, outperforming baselines by 63\% execution success while generalizing to unseen query poses and unseen object shapes.}
Project page:  \url{https://sites.google.com/view/neural-grasp-distance-fields}.
\end{abstract}

\section{Introduction}
\label{sec:intro}

\begin{figure}[t]
    \centering
    \begin{subfigure}[t]{0.4\columnwidth}
        \includegraphics[width=1.0\columnwidth, trim={0cm 4cm 0cm 3cm},clip]{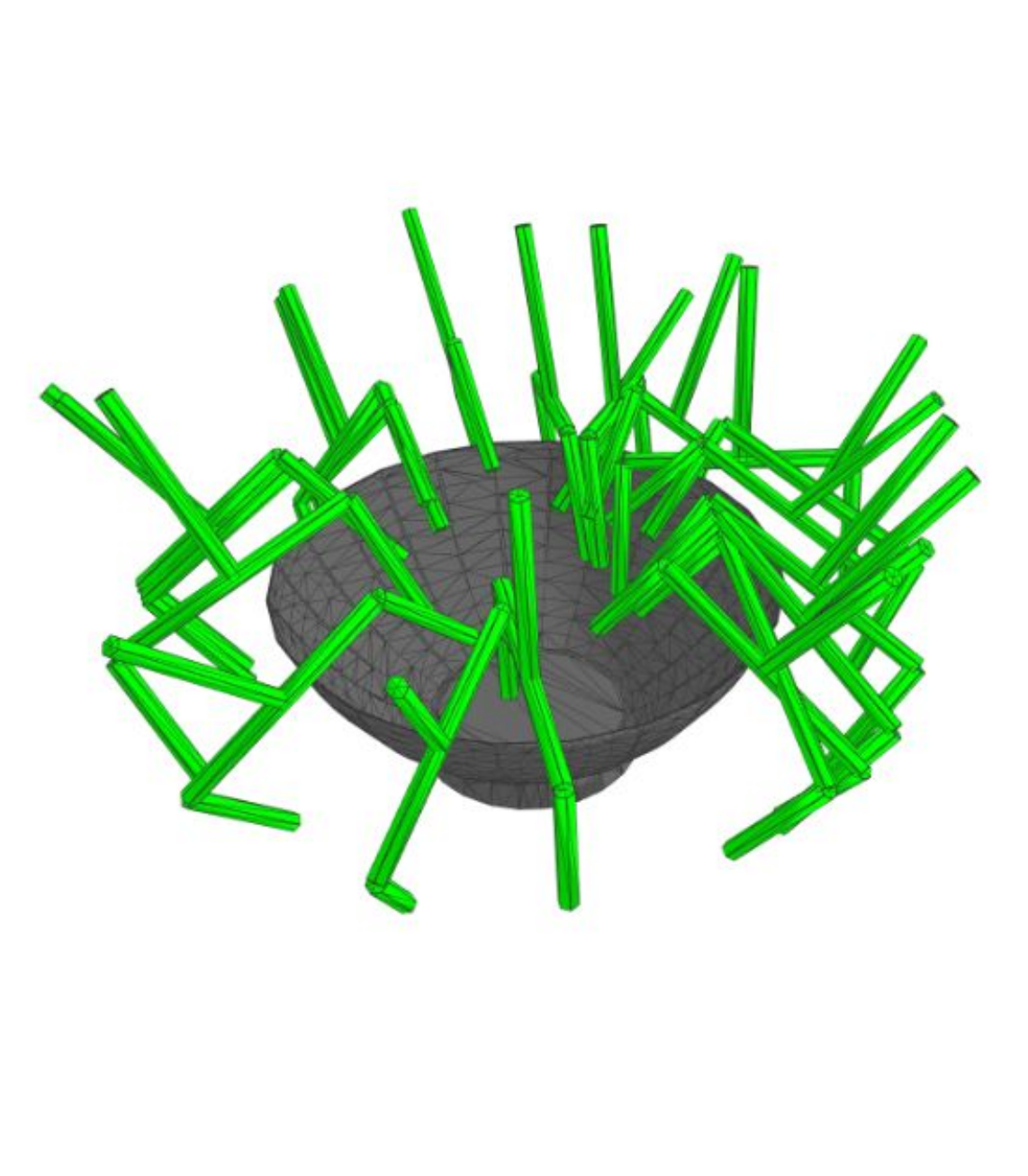}
        \caption{Discrete Grasp Set}
    \end{subfigure}
    \begin{subfigure}[t]{0.58\columnwidth}
        \includegraphics[width=1.0\columnwidth, trim={0cm 2cm 0cm 0cm},clip]{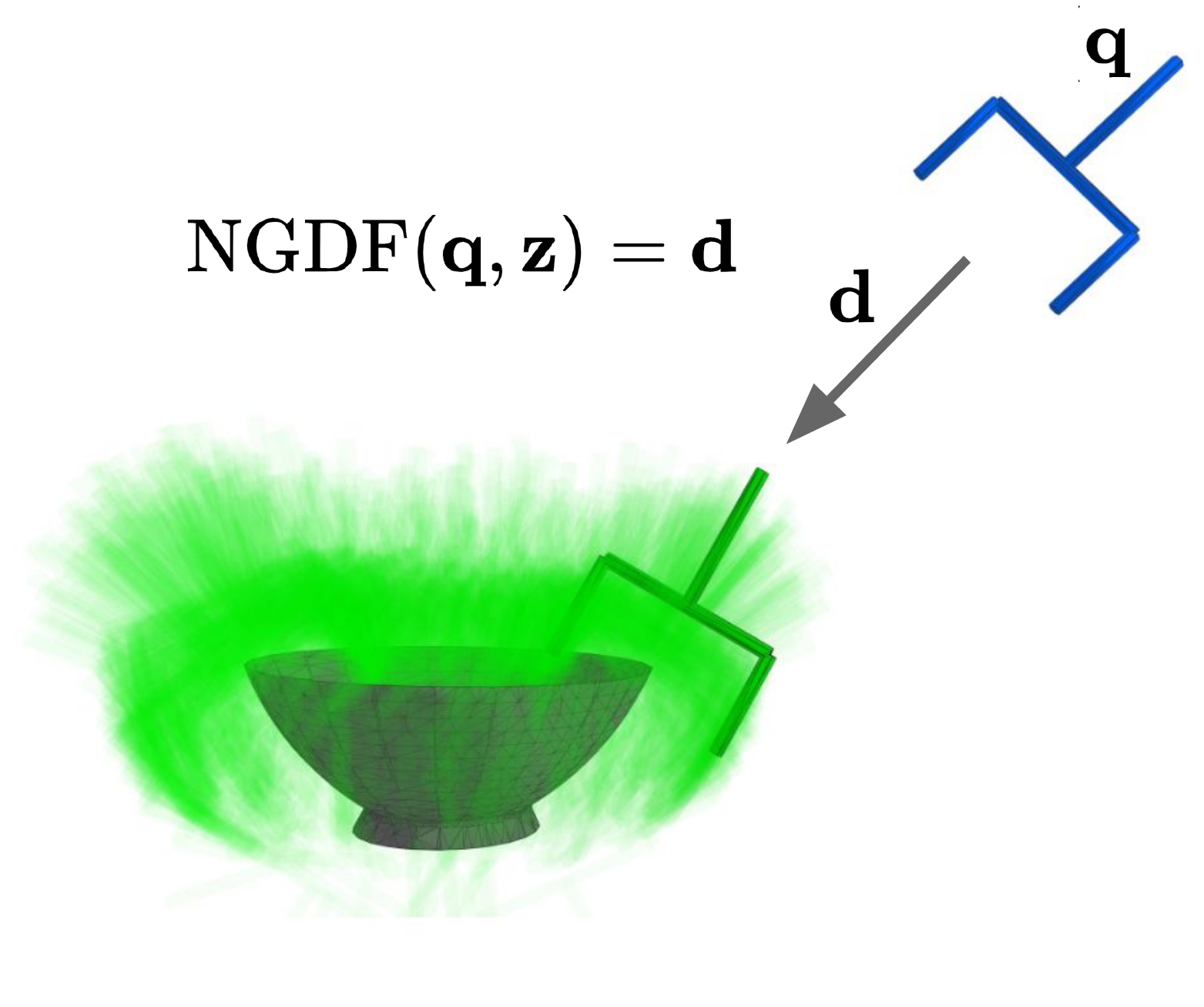}
        \caption{Continuous Grasp Manifold with \ngdf}
    \end{subfigure}
    \caption{
    (a) Existing grasp estimation methods produce discrete grasp sets which do not represent the true continuous manifold of possible grasps. (b) Our work, Neural Grasp Distance Fields (\ngdf), learns a continuous grasp manifold: given a query pose $\mathbf{q}$ and an object shape embedding $\mathbf{z}$, \ngdf outputs the distance $\mathbf{d}$ between $\mathbf{q}$ and the closest grasp. 
    This distance can be leveraged as a cost for optimization, facilitating joint grasp and motion planning.
    }
    \label{fig:teaser}
    \vspace{-1.5em}
\end{figure}

We present \textit{\textbf{N}eural \textbf{G}rasp \textbf{D}istance \textbf{F}ields} (\ngdf), which model the continuous manifold of valid grasp poses as the level set of a neural implicit function. 
Given a 6D query pose, \ngdf predicts the unsigned distance between the query and the closest valid grasp on the manifold (see \fig{fig:teaser}).

Neural implicit fields have driven recent advancements in novel view synthesis~\cite{mildenhall2020nerf} and 3D reconstruction~\cite{park2019deepsdf, mescheder2019occupancy, chibane2020ndf, chibane20ifnet}.
These approaches represent distributions as continuous functions that take a query as input and predict its relationship to the learned distribution. 
In 3D shape reconstruction, for instance, neural implicit fields are used to represent the surface of a shape: 3D points are used as queries, and the output is the distance to the surface, or occupancy at the query point. 
Unlike explicit methods, neural implicit fields can encode complex topological distributions and are not limited by resolution. 

With \ngdf, formulating grasp learning as a neural field allows us to interpret the implicit function as a cost such that a query pose can be optimized to result in a grasp pose.
Prior grasp estimation methods largely output a discrete set of candidate grasps~\cite{sundermeyer2021icra, mousavian20196, pointnet++grasping,liang2019pointnetgpd,fang2020graspnet}, from which one grasp must be selected to perform downstream planning. 
Instead, we incorporate the grasp distance cost directly into a gradient-based optimizer~\cite{zucker2013chomp} to jointly optimize the grasp and reaching motion from an initial trajectory.  
During each optimization iteration, \ngdf estimates the distance between the final gripper pose of the trajectory and the grasp level set.
This ``grasp distance'' is minimized as a cost, along with other trajectory costs such as smoothness and collision avoidance. 
The gradient of the grasp cost for updating the trajectory is computed through fully differentiable operations.
This optimization results  in a smooth, collision-free trajectory that reaches a valid grasp pose. 

\hl{In experiments, we find that \ngdf learns the level set of valid grasp poses, outperforms baselines by 63\% execution success on simulated reaching and grasping, and generalizes to unseen object shapes and poses in the real world. 
}
The key contributions of the paper are:
\begin{itemize}
\item Neural Grasp Distance Fields (\ngdf), a neural implicit function that 
\hl{predicts the distance between a query pose and the closest grasp, representing the manifold of grasps as a continuous level set.}
\item A gradient-based optimization algorithm that incorporates \ngdf for \hl{joint} reach and grasp planning.
\end{itemize}

\begin{figure*}[t]
\centering 

\usetikzlibrary{positioning}
\usetikzlibrary{calc}
\usetikzlibrary{arrows.meta}

\tikzset{
    header/.style = {font=\small, rectangle, minimum width=2.5cm, minimum height=0.75cm},
    box/.style = {font=\small, rectangle, draw=black!50, fill=black!5, thick, minimum width=2.3cm, minimum height=0.85cm},
    heuristic/.style = {font=\small, rectangle, draw=black!50, fill=yellow!25, thick, minimum width=2.2cm, minimum height=0.55cm},
}

\begin{tikzpicture}
\path 
    (2.75,0) node(h1) [header] {Perception}
    (6,0) node(h2) [header,minimum width=3.3cm,right=0.6cm of h1.east] {Grasp Estimation}
    (9.25,0) node(h3) [header,right=0.6cm of h2.east] {Grasp Selection}
    (12.15,0) node(h4) [header,right=0.6cm of h3.east] {Motion Planning};
\draw[thick,->] (node cs:name=h1) -- (node cs:name=h2);
\draw[thick,->] (node cs:name=h2) -- (node cs:name=h3);
\draw[thick,->] (node cs:name=h3) -- (node cs:name=h4);

\path 
    (3,-1.75) node(p1) [box,text width=2.35cm,align=center,draw,below=1.15cm of h1] 
        {Object Pose\\(Known Shape)}
    (3,-4.25) node(p2) [box,text width=2.35cm,align=center,draw,below=0.15cm of p1] 
        {Point Cloud\\(Unknown Shape)};
    
\path 
    (6,-1) node(e1) [box,align=center,draw,minimum width=3.3cm,below=0.15cm of h2] 
        {Discrete Set, Known}
    (6,-2) node(e2) [box,align=center,draw,below=0.15cm of e1,minimum width=3.3cm] 
        {Discrete Set, Predicted\\~\cite{sundermeyer2021icra,mousavian20196,pointnet++grasping,liang2019pointnetgpd,fang2020graspnet}}
    (6,-3) node(e3) [box,align=center,draw,below=0.15cm of e2,,minimum width=3.3cm] 
        {Continuous Set, Known}
    (9,-4) node(e4) [box,align=center,draw,below right=0.15cm and 0cm of e3.south west,fill=\ngdfclr!30,minimum width=6.35cm] 
        {Continuous Set, Predicted};
    
\path
    (9,-2.0) node(s2) [heuristic,align=center,draw,below=0.5cm of h3] 
        {Min. Distance}
    (9,-3.0) node(s3) [heuristic,align=center,draw,below=0.25cm of s2] 
        {Max Score}
    (9,-4.0) node(s4) [heuristic,align=center,draw,below=0.25cm of s3] 
        {Adaptive Cost};
    
\path 
    (10,-1.75) node(mp1) [box,align=center,draw,below=1.15cm of h4] 
        {Sampling\\~\cite{844730,508439}}
    (10,-4.25) node(mp2) [box,align=center,draw,below=0.15cm of mp1] 
        {Optimization\\~\cite{zucker2013chomp,mukadam2018continuous}};

\path 
    (10,-3) node(ge) [font=\small, rectangle, draw=black!50, fill=black!5,minimum width=1.8cm,minimum height=0.3cm, thick,rotate=-90,above right=0.97cm and 0.7cm of mp2] {Execution};

\path 
    (0,-1.5) node(m1) [font=\small,rectangle,rounded corners,draw,fill=\omgclr!30,align=center,minimum width=2cm] 
        {OMG~\cite{wang2020manipulation}}
    (0,-2.25) node(m2) [font=\small,rectangle,rounded corners,draw,fill=\cbrrtclr!30,align=center,minimum width=2cm,below=0.2cm of m1] 
        {CBiRRT~\cite{berenson2011task}}
    (0,-3) node(m4) [font=\small,rectangle,rounded corners,draw,fill=\bclr!30,align=center,minimum width=2cm,below=0.2cm of m2]
        {B1}
    (0,-3.75) node(m3)  [font=\small,rectangle,rounded corners,draw,fill=\ngdfclr!30,align=center,minimum width=2cm,below=0.2cm of m4]
        {\ngdf (Ours)};

\draw[line width=0.5mm,\omgclr!50,dotted]
    (m1.east) -- (p1.165)
    (p1.15) -- (e1.west)
    (e1.east) -- (s4.west)
    ;
\draw[-{Stealth[length=2.5mm, width=2.5mm]},line width=0.5mm,\omgclr!50,dotted]
    (s4.358) to[out=-45,in=-150] (mp2.178);
\draw[-{Stealth[length=2.5mm, width=2.5mm]},line width=0.5mm,\omgclr!50,dotted]
    (mp2.176) to[out=120,in=30] (s4.2);
\draw[-{Stealth[length=3mm, width=3mm]},line width=0.5mm,\omgclr!50,dotted]
    (mp2.2) -- (ge.332);
    
\draw[-{Stealth[length=3mm, width=3mm]},line width=0.5mm,\bclr!60,dashed]
    (m4.east) -- (p2.170)
    (p2.15) -- (e2.west)
    (e2.east) -- (s3.west)
    (s3.east) -- (mp2.165)
    (mp2.15) -- (ge.309)
;

\draw[-{Stealth[length=3mm, width=3mm]},line width=0.5mm,\cbrrtclr!50]
    (m2.east) -- (p1.191)
    (p1.349) -- (e3.west)
    (e3.east) -- (s2.west)
    (s2.east) -- (mp1.west)
    (mp1.east) -- (ge.204);

\draw[line width=0.5mm,\ngdfclr!50]
    (m3.east) -- (p2.190)
    (p2.350) -- (e4.180);
\draw[-{Stealth[length=2.5mm, width=2.5mm]},line width=0.5mm,\ngdfclr!50]
    (mp2.198) to[out=140,in=60] (e4.8);
\draw[-{Stealth[length=2.5mm, width=2.5mm]},line width=0.5mm,\ngdfclr!50]
    (e4.7) to[out=-30,in=-120] (mp2.200);
\draw[-{Stealth[length=3mm, width=3mm]},line width=0.5mm,\ngdfclr!50]
    (mp2.348) -- (ge.342);

\end{tikzpicture}
\caption{Columns illustrate design decisions within grasp and motion planning pipelines.
The left-most column highlights representative pipelines like {\setlength{\fboxsep}{1pt}\colorbox{\omgclr!30}{OMG-Planner}}~\cite{wang2020manipulation}, {\setlength{\fboxsep}{1pt}\colorbox{\cbrrtclr!30}{CBiRRT}}~\cite{berenson2011task}, and baseline {\setlength{\fboxsep}{1pt}\colorbox{\bclr!30}{B1}} from Table~\ref{table:eval-reach-grasp} which uses a SOTA grasp estimator~\cite{sundermeyer2021icra}.
The respective design choices for these methods are traced through the columns.
We identify {\setlength{\fboxsep}{1pt}\colorbox{\ngdfclr!30}{learned continuous representations}} as an under-explored option for grasp estimation, and propose {\setlength{\fboxsep}{1pt}\colorbox{\ngdfclr!30}{\ngdf}} as a solution that does not require a heuristic {\setlength{\fboxsep}{1pt}\colorbox{yellow!50}{grasp selection step}} since the grasp pose is jointly optimized with motion planning. 
}
\label{fig:related}
\vspace{-1.5em}
\end{figure*}
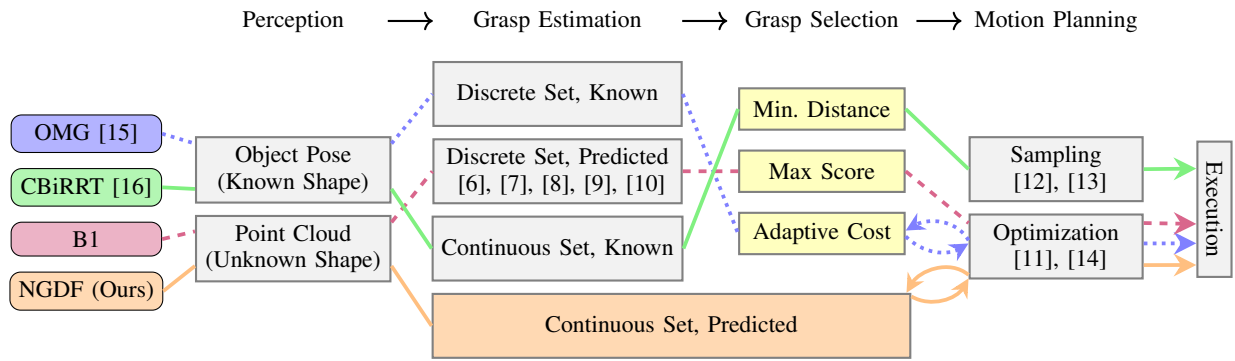

\section{Related Work}
\label{sec:related}
 
While grasping and motion planning are well-studied topics in robotics, prior works often propose \hl{different system designs with different assumptions}, making comparison and contextualization difficult. 
We \hl{summarize} the most important design decisions for 6-DOF grasp and motion planning and trace the decisions \hl{in representative methods (see \fig{fig:related})}. 

\subsection{6-DOF Grasp Estimation}
\label{sec:6dgrasp}

\hl{
6-DOF grasp estimation is a well-studied task~\cite{6672028, kleeberger2020survey} that aims to predict successful grasps in SE(3) for target objects; we focus here on recent, data-driven methods.
State-of-the-art methods take point clouds as input and output a discrete set of grasps, representing only a subset of the true continuous grasp set~\cite{mousavian20196, sundermeyer2021icra, pointnet++grasping, liang2019pointnetgpd,fang2020graspnet}. 
}
Outputting a finer discretization comes with a cost of a greater computational complexity for both grasp estimation as well as grasp selection: a final grasp must be chosen from the predicted set.
Because these methods only predict discrete grasp sets, they necessitate a multi-stage approach, which can be brittle if any of the stages (grasp estimation, selection, or motion planning) fails. 
\hl{Our single-stage approach models grasps as the level set of a continuous implicit function to jointly optimize grasping and motion planning.}

\subsection{Joint Grasp Selection and Motion Planning}
\label{sec:rel-jgmp}
 
 \hl{
Following the multi-stage paradigm above, several works assume a grasp set is provided by an upstream method, and address the downstream task of planning a reaching trajectory.
Berenson~\etal\cite{berenson2011task} model grasp sets as a continuous range of poses called Task Space Regions, and use sampling-based planning to satisfy the constraint.
GOMP~\cite{ichnowski2020gomp}, uses sequential quadratic programming on discrete grasp sets for fast bin picking.
Goal-set CHOMP~\cite{goalsetchomp} incorporates hard constraints like goal sets into trajectory optimization.
The methods above do not address the problem of switching between grasps during planning;
OMG-Planner~\cite{wang2020manipulation} therefore proposes online learning to estimate goal costs and switch to the minimum cost grasp at every optimization iteration.
OMG-Planner used ground-truth grasp sets per object, though their method can use estimated grasp sets as well.
Our approach does not assume grasps are provided and does not require explicit grasp selection; instead, \ngdf estimates and updates the grasp pose during trajectory optimization itself.
}

\hl{
Other works propose closed-loop methods for 6-DOF grasping.
}
Wang~\etal\hl{\cite{wang2022latent}}
learn a latent space of trajectories for closed-loop grasping.
\hl{
Song~\etal\cite{song2020grasping} learn a closed-loop policy from human demonstrations.
Temporal GraspNet~\cite{yang2020reactive} updates a discrete grasp set over time by querying a grasp evaluator. 
In this work, we introduce a novel implicit representation for the grasp manifold. 
We focus on open-loop planning and leave closed-loop planning with \ngdf as future work.
}

\subsection{Implicit Neural Representations}
\label{sec:rel-inr}
\hl{
Recent advances in vision and graphics research have used implicit neural representations to achieve impressive results on novel view synthesis~\cite{mildenhall2020nerf} and 3D reconstruction~\cite{mescheder2019occupancy, chibane20ifnet, park2019deepsdf, chibane2020ndf}.
}
Karunratakul~\etal\cite{karunratanakul2020grasping} learn an implicit representation for human grasp poses.
\hl{
Inspired by these works, we learn an implicit neural function to predict distances between query gripper poses and grasp poses, and use this function to optimize grasp trajectories. 
}

\hl{
The robotics community has also explored neural implicit functions for a variety of manipulation tasks~\cite{zhu2021rgb, IchnowskiAvigal2021DexNeRF, wi2022virdo, florence2022implicit, simeonovdu2021ndf, li20223d, driess2022learning}.}
GIGA~\cite{jiang2021synergies} proposed using neural implicit functions to model both 3D shape and grasp quality. 
\hl{
However, GIGA predicts a single grasp parameterization per 3D location, and requires a sampling procedure to select the final pose from the implicit set.
}
Our approach predicts grasp distance, allowing multiple grasp orientations per 3D location, and uses optimization to minimize grasp distance and achieve the grasp pose.

Concurrent works have proposed continuous representations for dexterous hands~\cite{wu2022ILAD} and multiple grippers~\cite{khargonkar2022neuralgrasps}. 
Urain~\etal\cite{urain2022se} represents grasps as diffusion fields, framing joint grasp and motion planning as an inverse diffusion process. 
In this paper, we use an implicit function to represent grasp distance, and use gradient-based trajectory optimization for joint grasp and motion planning. 

\begin{figure*}[th]
    \centering
    \includegraphics[width=0.95\textwidth]{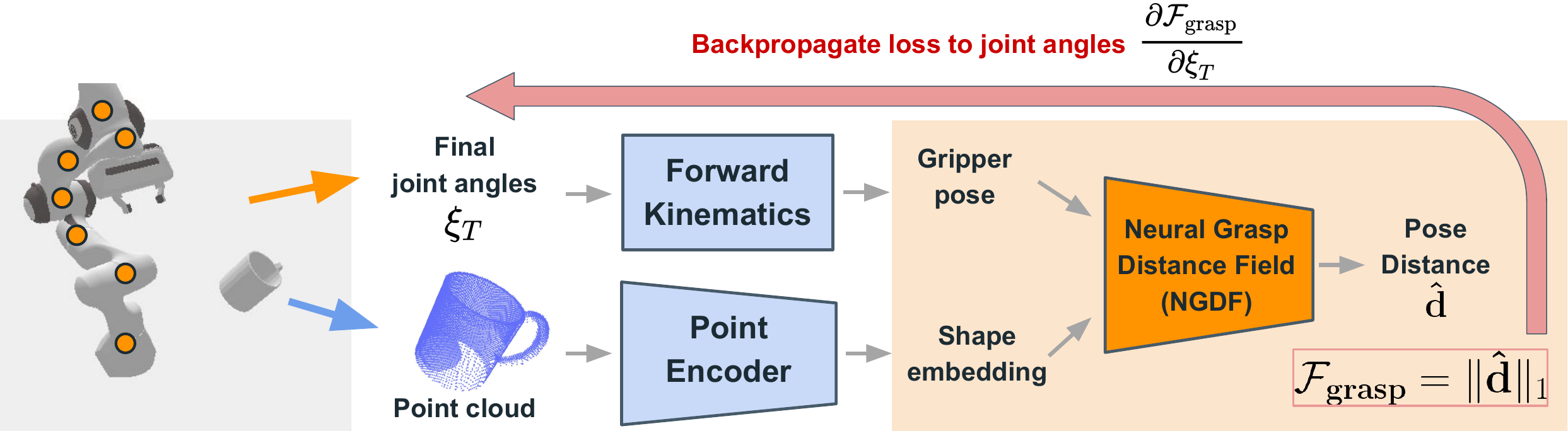}
    \caption{We use \ngdf as a goal cost function on the final state of a trajectory during gradient-based optimization. Given the current robot joint configuration and a point cloud of an object or scene, the current gripper pose and a shape embedding are computed as inputs for \ngdf. Then, \ngdf predicts the distance of the current gripper pose to the closest grasp (\sect{sec:ngf}). The predicted distance is used as the cost and the gradient with respect to the joint configuration is computed with backpropagation. This cost (with gradient) is used with other costs like smoothness and collision avoidance to update the trajectory (\sect{sec:jgmp}).}
    \label{fig:approach}
    \vspace{-1.5em}
\end{figure*}

\section{Background}
\label{sec:background}

\textbf{Neural Implicit Functions}.
Neural implicit functions (NIFs) are neural networks that take a query $\mathbf{q} \in \mathbb{R}^d$ and optionally a context embedding $\mathbf{z} \in \mathcal{Z}$ to output a scalar value that represents a relationship to an underlying distribution: $f(\mathbf{q}, \mathbf{z}): \mathbb{R}^d \times \mathcal{Z} \mapsto \mathbb{R}$. 
In the domain of 3D shape reconstruction, the context $\mathbf{z}$ is a latent shape embedding, the query $\mathbf{q}$ is a 3D point, and the scalar output is either distance to the closest surface~\cite{park2019deepsdf,chibane2020ndf}, or occupancy~\cite{mescheder2019occupancy,chibane20ifnet}.
The shape surface is represented by the zero level set in distance-based methods, or the decision boundary in occupancy-based methods.
Unlike explicit functions, NIFs are not limited by resolution as they predict a value at any query point,
and also better represent underlying distributions that are disjoint~\cite{florence2022implicit}.
Our approach leverages both properties in learning a manifold of grasps.

\textbf{Gradient-based Trajectory Optimization}.
\hl{
A mapping from time $t$ to robot joint configuration $\mathbf{p}$ is defined as a trajectory $\xi: [0, T] \rightarrow \mathbf{p}$.
Trajectory optimization aims to find the optimal trajectory given an objective functional $\mathcal{U}$:
}
\begin{equation}
    \textstyle
    \xi^* = \arg \min_{\xi} \mathcal{U}[\xi],\text{ s.t. }\xi(0) = \mathcolorbox{yellow}{\mathbf{p}_s},\ \xi(T) = \mathcolorbox{yellow}{\mathbf{p}_g}
\end{equation}
\hl{for a given start $\mathbf{p}_s$ and goal $\mathbf{p}_g$ configuration}.
In manipulation, the objective $\mathcal{U}$ contains cost terms for smoothness and collision avoidance. 
CHOMP~\cite{zucker2013chomp} solves for $\xi^*$ with functional gradient descent: 
\begin{equation}
\label{eq:chomp-update}
    \xi_{t+1} = \xi_t - \eta A^{-1} \bar{\nabla} \mathcal{U}(\xi_t)
\end{equation}
where $A$ is an acceleration metric that helps propagate updates over the entire trajectory.

\section{Method}
\label{sec:approach}

In this work, we represent a set of poses $\mathcal{M} \subset \mathbf{SE}(3)$ as the level set of a neural implicit function.
This implicit function takes a query pose $\mathbf{q}$ as input and estimates its distance to the learned level set.
\sect{sec:ngf} describes how Neural Grasp Distance Fields (\ngdf) leverage this insight to learn the level set of valid \textit{grasp} poses.
\sect{sec:jgmp} explains how to incorporate \ngdf into a trajectory optimization framework to jointly reason over smooth and collision-free reaching trajectories that end at a valid grasp pose.
\fig{fig:approach} provides an overview of our method.

\subsection{Neural Grasp Distance Fields}
\label{sec:ngf}

Given a query pose $\textbf{q} \in \mathbf{SE}(3)$ and a shape embedding $\mathbf{z} \in \mathcal{Z}$, \ngdf defines an implicit function: 
$
    \text{\ngdf}(\mathbf{q}, \mathbf{z}) = \mathbf{d}
$,
where $\mathbf{d}$ is the distance from $\mathbf{q}$ to the closest valid grasp $\mathbf{g} \in \mathcal{M} \subset \mathbf{SE}(3)$ for an object in a scene.
Valid grasps are poses where a gripper can stably grasp an object by closing its fingers.
For the distance metric $\mathbf{d}$ we combine translation and orientation distances into a single ``control points'' metric~\cite{mousavian20196}:
\begin{equation}
\label{eq:ctrl_pts}
    d_i = \| \mathcal{T}(\mathbf{q}; \mathcolorbox{yellow}{\mathbf{c}_i}) - \mathcal{T}(\mathbf{g}; \mathcolorbox{yellow}{\mathbf{c}_i}) \|_1\ ,\ i = 0, \dots, N
\end{equation}
\noindent where $\mathcal{T}(\cdot; \mathcolorbox{yellow}{\mathbf{c}_i})$ is the transformation of a predefined set of points $\{\mathcolorbox{yellow}{\mathbf{c}_i}\}$ on the gripper. 
Since $\mathbf{q}$ and $\mathbf{g}$ belong to $\mathbf{SE}(3)$, the distance could be defined based on the manifold geodesic distance between those poses, however we find that the control points based distance metric balances the translation and rotation costs better in practice.
\ngdf estimates the distance for each control point $\mathcolorbox{yellow}{\mathbf{c}_{0 \dots N}}$ separately:
$\mathbf{d}(\mathbf{q}, \mathbf{g}) = \begin{bmatrix}
        d_0,
        \dots,
        d_N
    \end{bmatrix}^T$.
During training, the estimated distances $\hat{\mathbf{d}}$ are supervised with L1 loss: $\mathcal{L} = \| \hat{\mathbf{d}} - \mathbf{d} \|_1$.

\subsection{Optimization of Grasping Trajectories using \ngdf}
\label{sec:jgmp}

For a given query pose, \ngdf outputs the distance to the closest grasp pose. 
We now show how to enable joint optimization for reaching and grasping with NGDF. 
We incorporate \ngdf as a goal cost estimator within a gradient-based trajectory optimizer that already has cost terms for smoothness and collision avoidance.

In this work, we combine \ngdf with CHOMP~\cite{zucker2013chomp} (described in  \sect{sec:background}), though \ngdf can be used in any gradient-based trajectory optimization algorithm. 
Since CHOMP specifies a fixed goal $\mathcolorbox{yellow}{\mathbf{p}_g}$, we modify CHOMP to include $\mathcolorbox{yellow}{\mathbf{p}_g}$ as a variable in the optimization following Dragan~\etal\cite{goalsetchomp}.
We then add our grasp cost $\mathcal{F}_{\rm grasp}$ as the variable goal cost to the objective functional $\mathcal{U}$:
\begin{equation}
\label{eq:objfunc}
    \mathcal{U}[\xi] =
        \lambda_1 \bm{\mathcal{F}_{\rm grasp}[\xi]} +
        \lambda_2 \mathcal{F_{\rm smooth}}[\xi] + \lambda_3 \mathcal{F_{\rm obs}}[\xi]
\end{equation}
where $\lambda_i$ are cost weights.

\textbf{Grasp Distance as a Goal Cost}.
We now define $\mathcal{F}_{\rm grasp}$ and derive its functional gradient $\bar{\nabla}\mathcal{F}_{\rm grasp}$ for gradient-based optimization.
For a trajectory (during any iteration of optimization), we calculate the gripper pose from the final joint configuration using forward kinematics: $\mathbf{q}_T = {\rm FK(\xi_T)}$.
\hl{We then use \ngdf to estimate the distance of this gripper pose to a valid grasp: ${\rm \ngdf}(\mathbf{q}_T, \mathbf{z}) = \hat{\mathbf{d}}$. The norm of this distance becomes our grasp cost: $\mathcal{F_{\rm grasp}}[\xi] = \| \hat{\mathbf{d}} \|_1$.}
We can compute the gradient of the grasp cost with respect to the joint configuration $\xi_T$ through backpropagation:
\begin{equation}
\label{eq:backprop}
    \frac{\partial \mathcal{F_{\rm grasp}}}{\partial \xi_T} = 
        \mathcolorbox{yellow}{
        \frac{\partial \mathcal{F_{\rm grasp}}}{\partial 
        \mathbf{q}_T}
        \frac{\partial \mathbf{q}_T}{\partial \rm FK}
        }
        \frac{\partial \rm FK}{\partial \xi_T}
\end{equation}

Since the grasp cost only applies to the final configuration in a trajectory, the functional gradient $\bar{\nabla}\mathcal{F}_{\rm grasp}$ contains all zeros except for the last row: $\bar{\nabla}\mathcal{F}_{\rm grasp} = [\mathbf{0}, \mathbf{0}, \dots, \frac{\partial \mathcal{F}_{\rm grasp}}{\partial \xi_T}]^T$. 
\vspace{0.2em}

\textbf{Joint Optimization of Trajectory Costs}.
Similar to the objective functional (Eq.~\ref{eq:objfunc}), the objective functional gradient $\bar{\nabla} \mathcal{U}$ is a weighted sum of gradients:
$
    \bar{\nabla} \mathcal{U}[\xi] =
        \lambda_1 \bar{\nabla} \mathcal{F_{\rm grasp}} +
        \lambda_2 \bar{\nabla} \mathcal{F_{\rm smooth}} + \lambda_3 \bar{\nabla} \mathcal{F_{\rm obs}}.
$
At every optimization iteration, we compute the costs and functional gradients as described above, then update the trajectory according to the $A$-metric update rule (Eq.~\ref{eq:chomp-update}).
Since our objective cost has terms for minimizing distance to a valid grasp, maintaining smoothness, and avoiding collisions, our algorithm jointly optimizes all three to produce reaching and grasping trajectories.

\subsection{Implementation Details}
\label{sec:impl-details}

\textbf{Dataset}. Training \ngdf requires a dataset of point clouds, valid grasp poses, and query poses.
We use the ACRONYM~\cite{acronym2020} dataset, which contains object meshes and successful grasp poses collected in NVIDIA FleX~\cite{10.1145/2601097.2601152}.
For grasp poses, our evaluations in \sect{sec:results} are run in PyBullet~\cite{coumans2021}, so we relabel the successful grasp poses based on their success in PyBullet with the same linear and rotational shaking parameters used in ACRONYM.
In addition, we filter the positive grasp set to only include grasps
where the normals at the mesh and finger contact points are opposed to each other (-0.98 cosine similarity).
Our results in \sect{sec:eval-gmp} show that this filtering improves grasp performance. 
To collect query poses for the dataset, we sample 1 million random $\mathbf{SE}(3)$ poses within a \SI{0.5}{\meter} radius of the object mesh centroid.
\hl{While it is possible that some of the sampled poses could be positive grasps, we assume they are few in number and do not run additional grasp evaluation to filter them.}
For each sampled pose, we use distance to the closest grasp in the valid grasp set (see \sect{sec:ngf}) as our supervision.

\textbf{Architecture}. 
An input point cloud is converted into the shape embedding $\mathbf{z}$ using a VN-OccNet~\cite{deng2021vector} encoder pre-trained on 3D reconstruction~\cite{simeonovdu2021ndf}.
The input to \ngdf is a concatenation of this shape embedding $\mathbf{z}$ with the input query $\mathbf{q}$'s position and quaternion. 
The \ngdf network is based on DeepSDF~\cite{park2019deepsdf} and consists of 8 MLP layers, 512 units each, and ReLU activations on the hidden layers.
A softplus activation on the output layer ensures positive outputs.

\textbf{Training Procedure}.
We freeze the weights of the pre-trained point encoder during training and only train the \ngdf network. 
\hl{Each training sample consists of a partial point cloud, a query pose, and the closest valid grasp.}
\hl{Similar to NDF~\cite{simeonovdu2021ndf},} the partial point cloud is merged together from 4 camera views and downsampled to 1500 points using farthest point sampling. 
Random rotation augmentations are applied to \hl{each sample with 70\% probability}.
Finding the ground truth closest grasp pose is computationally expensive and requires multiple simulated grasp attempts per query pose. 
Therefore, our supervision is pseudo-ground truth, as the closest grasp pose comes from a large but discrete set of grasps~\cite{acronym2020}.
We find that this discrete grasp set is dense enough to train \ngdf, while still representing unseen valid grasp poses at or near the zero level set (\sect{sec:eval-ngf}).

\textbf{Trajectory Optimization}.
CHOMP~\cite{zucker2013chomp} uses a fixed or decaying step size for functional gradient updates, which is sufficient for trajectories with fixed start and goal joint configurations. 
However, with our modification of CHOMP in~\sect{sec:jgmp} to allow a variable goal configuration, we found that such simple step size strategies resulted in poor convergence. 
We address this issue by using Adam~\cite{Kingma2015AdamAM} to adaptively update the step size (``CHOMP-Adam'').
We use differentiable SE(3) operations~\cite{pineda2022theseus} and a differentiable robot model~\cite{pmlr-v120-sutanto20a} to backpropagate gradients from the output of \ngdf to the robot joint configuration (Eq.~\ref{eq:backprop}).

\section{Experiments}
\label{sec:results}

We first evaluate how well \ngdfs represent valid grasp manifolds as their zero level sets (\sect{sec:eval-ngf}). 
Then we perform a full system evaluation with \ngdfs on a ``reaching and grasping" task (\sect{sec:eval-gmp}), where an \ngdf is used within a gradient-based trajectory optimizer as a goal cost function.
We evaluate generalization on grasping intra-category unseen objects (\sect{sec:intra-category}), and demonstrate grasping on a real robot system (\sect{sec:real}).

\subsection{\hl{\ngdf Level Set Evaluation}}
\label{sec:eval-ngf}

First, we investigate whether the learned level set of an \ngdf represents successful grasps.
Our evaluation procedure considers driving an initial query pose to the learned level set. 
We use the distance output from \ngdf as a loss, and update the query pose with Adam~\cite{Kingma2015AdamAM} using backpropagated gradients. 
Note that this evaluation optimizes just the gripper pose; full-arm trajectory optimization is considered in the next subsection.
We evaluate \ngdf on three objects: Bottle, Bowl, and Mug.
For this evaluation, we train a single \ngdf model for each object, and evaluate models trained with and without the dataset filtering procedure described in~\sect{sec:impl-details}.
We run the optimization for 3k steps with a learning rate of 1e-4. 
Since we represent poses as positions and quaternions, we normalize the quaternion after each gradient update to ensure valid rotations.

\begin{figure}[!t]
    \centering
    \includegraphics[width=0.9\columnwidth]{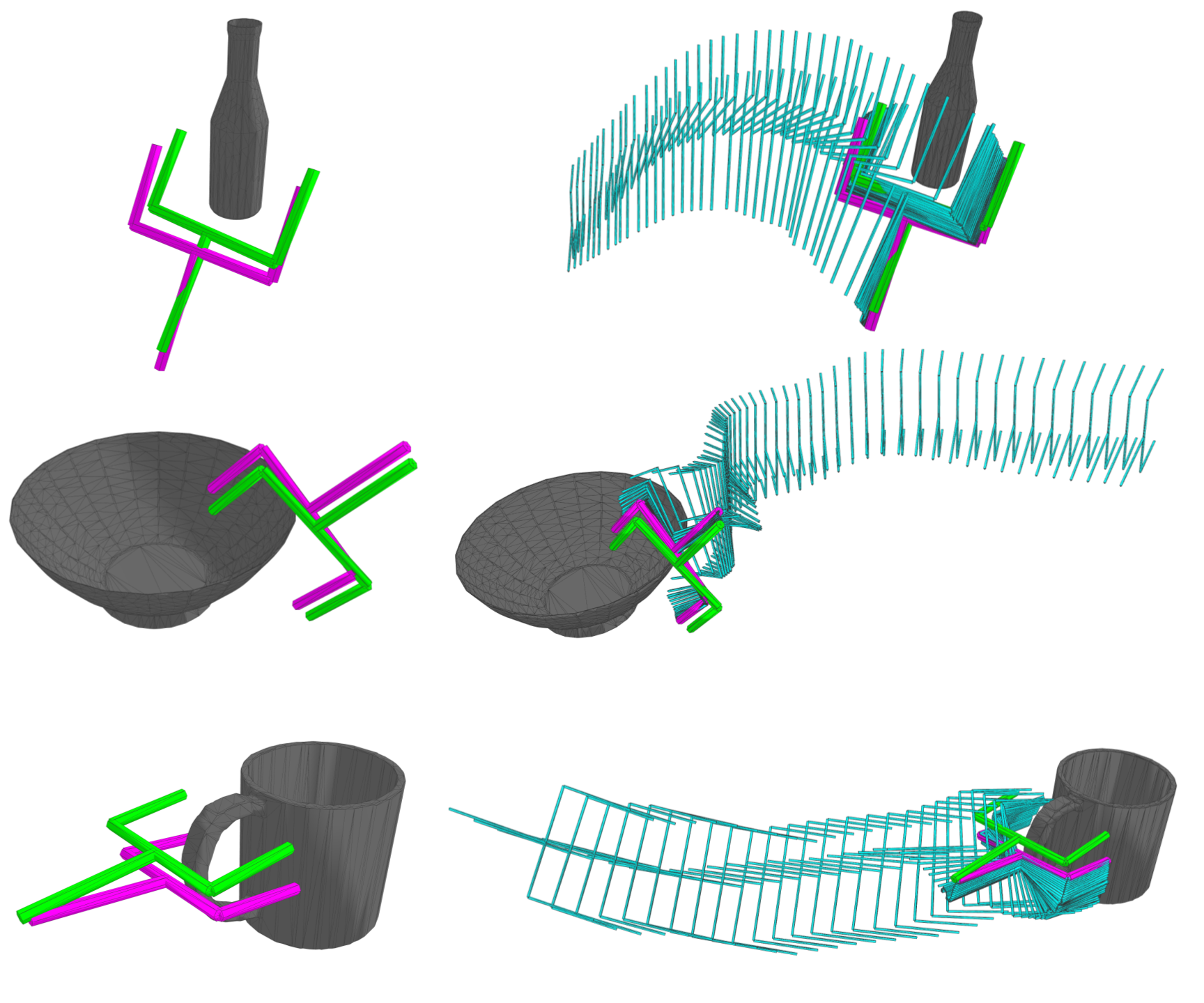}
    \caption{Grasp Level Set Evaluation. Left: Final predicted pose (\textcolor{magenta}{magenta}) and its closest grasp pose (\textcolor{ForestGreen}{green}) in the training dataset. Right: Gripper path (\textcolor{teal!80}{teal}) as it is optimized from initial to final pose.
    Object meshes shown for visual clarity; our method takes point clouds as input.}
    \label{fig:level_set}
    \vspace{-1em}
\end{figure}

\begin{table}[!t]
    \centering
    \caption{\normalsize \ngdf Grasp Level Set Results}
    \vspace{-0.5em}
    \label{table:level-set}
    \small
    \begin{tabular}{l c c}
        \toprule
        & Train Set Error (m) $\downarrow$
        & \hl{Grasp Success} $\uparrow$
        \\
        \midrule
        Bottle-NoFilter
            & $\mathbf{0.023 \pm 0.01}$ 
            & $0.480$
            \sd{\hl{$(0.34, 0.63)$}}
            \\
        Bottle
            & $0.029 \pm 0.01$ 
            & $\mathbf{0.880}$
            \sd{\hl{$(0.76, 0.95)$}}
            \\
        \midrule
        Bowl-NoFilter
            & $0.036 \pm 0.02$ 
            & $0.540$
            \sd{\hl{$(0.39, 0.68)$}}
            \\
        Bowl
            & $\mathbf{0.033 \pm 0.01}$ 
            & $\mathbf{0.760}$
            \sd{\hl{$(0.62, 0.87)$}}
            \\
        \midrule
        Mug-NoFilter
            & $0.038 \pm 0.01$
            & $0.680$
            \sd{\hl{$(0.53, 0.80)$}}
            \\
        Mug
            & $\mathbf{0.035 \pm 0.01}$ 
            & $\mathbf{0.860}$
            \sd{\hl{$(0.73, 0.94)$}}
            \\
        \bottomrule
    \end{tabular}
    \vspace{0.3em}
    \caption*{
    Results are averaged over 50 unseen query poses per object, sampled from within a \SI{0.5}{\meter} radius of the object centroid.
    }
    \vspace{-2.5em}
\end{table}

The quantitative results on grasp level set optimization are shown in Table~\ref{table:level-set}.
We use two metrics for this evaluation. 
The ``Train Set Error'' metric is the minimum control points distance (Eq.~\ref{eq:ctrl_pts}) between the optimized gripper pose and the closest grasp pose in the discrete training set.
Since \ngdf should learn a continuous level set and interpolate between grasps in the training set, we expect \ngdf not to achieve zero error on this metric, but it provides a good surrogate for comparing models.
The ``Grasp Success'' metric measures the grasp quality of the optimized gripper poses.
For each pose, we load the target object in PyBullet~\cite{coumans2021} and attempt a grasp at the specified pose.
The robot gripper is always initialized to the same position; the object is transformed relative to the gripper. 
Linear and rotational shaking are applied after gripping the object~\cite{acronym2020}, and the grasp is successful if the object is still gripped after the shaking.

\begin{table*}[t!]
    \centering
    \caption{\normalsize Reaching and Grasping Results}
    \vspace{-0.5em}
    \label{table:eval-reach-grasp}
    \small
    \begin{tabular}{c | l l l l | c }
        \toprule
        Method
        & \multicolumn{1}{c}{Perception} 
        & \multicolumn{1}{c}{Grasp Estimation}
        & \multicolumn{1}{c}{Grasp Selection} 
        & \multicolumn{1}{c|}{Goal}
        & \hl{Execution Success} $\uparrow$
        \\
        \midrule
        O1 \textit{(Oracle)}
            & \textit{Known Object Pose}
            & \textit{Known Discrete Grasps}
            & \textit{Min. Distance}
            & \textit{Fixed}
            & $0.96$
            \sd{\hl{$(0.89, 0.99)$}}
            \\ 
        {\setlength{\fboxsep}{1pt}\colorbox{\omgclr!30}{OMG~\cite{wang2020manipulation}}} \textit{(Oracle)}  
            & \textit{Known Object Pose}
            & \textit{Known Discrete Grasps}
            & \textit{Adaptive Cost}
            & \textit{Variable}
            & $\mathbf{0.99}$
            \sd{\hl{$\mathbf{(0.94, 1.00)}$}}
            \\ 
        \midrule
        {\setlength{\fboxsep}{1pt}\colorbox{\bclr!30}{B1}}
            & Unknown Object Pose
            & Predicted Discrete Grasps~\cite{sundermeyer2021icra}
            & Max Score
            & Fixed
            & $0.37$
            \sd{\hl{$(0.27, 0.47)$}}
            \\
        B2
            & Unknown Object Pose 
            & Predicted Discrete Grasps~\cite{sundermeyer2021icra}
            & Min. Distance
            & Fixed
            & $0.39$
            \sd{\hl{$(0.29, 0.50)$}}
            \\
        B3 
            & Unknown Object Pose
            & Predicted Discrete Grasps~\cite{sundermeyer2021icra}
            & Min. Distance
            & Variable
            & $0.38$
            \sd{\hl{$(0.28, 0.49)$}}
            \\
        B4
            & Unknown Object Pose
            & Predicted Discrete Grasps~\cite{sundermeyer2021icra}
            & Adaptive Cost
            & Variable
            & $0.31$
            \sd{\hl{$(0.22, 0.42)$}}
            \\ 
        {\setlength{\fboxsep}{1pt}\colorbox{\ngdfclr!30}{\ngdf (Ours)}}
            & Unknown Object Pose
            & Predicted Continuous Grasps
            & \textit{N/A} 
            & Variable
            & $\mathbf{0.61}$
            \sd{\hl{$\mathbf{(0.50, 0.71)}$}}
            \\
        \bottomrule
    \end{tabular}
    \vspace{0.1em}
    \caption*{
    Middle columns correspond to design decisions found in~\fig{fig:related}; color-coded methods also correspond to those shown in the same figure.
    }
\vspace{-2em}
\end{table*}

Our results show that while \ngdfs trained on filtered and unfiltered data have similar Train Set Error, the Grasp Success for filtered data models is much higher. 
These results also indicate that \ngdfs have learned continuous level sets, since the mean distance predicted by \ngdf after optimization is less than 1e-5, much lower than the minimum distance to the training set of grasps. 
\hl{\fig{fig:level_set} shows examples of the optimization path and achieved gripper pose.} 

\vspace{-0.25em}
\subsection{\hl{Simulated} Reaching and Grasping Evaluation}
\label{sec:eval-gmp}

Next, we evaluate our method on a full reaching and grasping task, which requires planning a smooth, collision-free grasping trajectory for the full robot arm starting from an initial robot joint configuration. This evaluates the full pipeline as opposed to just the stand-alone gripper pose in the previous subsection.
The task is considered successful if the robot executes the trajectory, closes its fingers to grasp the object, and lifts the object without losing it.
We place Bottle, Bowl, and Mug objects in simulation in 30 random orientations each \hl{\appref{(see~\appfig{fig:sim_traj}~\arxivref{in~\cite{weng2022neural},} left-most column)}}, thus 90 trials in total.
Our results indicate that even in a seemingly simple setting, randomly oriented objects present an overall challenging benchmark.

For this evaluation, we train a separate \ngdf (similar to NeRF approaches~\cite{mildenhall2020nerf, IchnowskiAvigal2021DexNeRF}) for each object, though our method can be extended to generalize across objects like other shape-conditioned implicit approaches~\cite{park2019deepsdf}. 
We also evaluate intra-category (known class, unseen shape) generalization in the next subsection. 
We run 500 iterations of CHOMP-Adam (see \sect{sec:impl-details}) with a learning rate of 3e-3. 
The grasp cost is weighted heavily relative to the collision and smoothness costs.
The trajectory is initialized using inverse kinematics so the gripper pose of the final joint configuration is within \SI{0.3}{\meter} of the center of the object point cloud; the rest of the initial trajectory is interpolated between the start and end joint configurations.

The results are shown in Table~\ref{table:eval-reach-grasp}.
We compare against oracle methods that provide upper-bound task performance, and against baselines that predict discrete grasps.
Oracle methods assume perfect object pose estimation and known discrete grasp set.
All discrete grasp methods run inverse kinematics over all discrete grasp goals and discard infeasible grasps.
\hl{For planning, methods use goal-set CHOMP~\cite{goalsetchomp} or CHOMP~\cite{zucker2013chomp}, depending on whether the goal is fixed or can vary.}
``O1'' selects the goal with minimum distance to the initial joint configuration, and keeps it fixed throughout planning.
``OMG''~\cite{wang2020manipulation} adaptively learns a cost for each grasp and selects the grasp with minimum cost at every optimization iteration (Variable Goal). 

The baselines that predict discrete grasps use Contact-Graspnet~\cite{sundermeyer2021icra} as the grasp estimator.
We use weights (provided by the authors) that are trained on millions of grasps and shapes.
``B1'' selects the grasp goal with the maximum score estimated by Contact-GraspNet and keeps it fixed during planning.
``B2'' selects the grasp goal with minimum distance to the initial joints and keeps it fixed during planning.
``B3'' allows varying grasps during planning using the minimum distance metric. 
``B4'' uses the same adaptive cost from OMG~\cite{sundermeyer2021icra} to select grasp goals during planning.

Our results show that while oracle methods perform well, methods that don't assume known object pose and use predicted grasps have much lower Execution Success.
Of the predicted grasp methods, \ngdf performs best.
Surprisingly, the B3 and B4 variable goal variants do not outperform fixed goal variants B1 and B2.
Failure cases for all methods are largely due to collisions between the gripper fingers and the object, which are a relatively small obstacle cost and may be difficult for the planner to balance with the other costs. 
\hl{\appref{\appfig{fig:sim_traj}~\arxivref{in~\cite{weng2022neural}} contains qualitative \ngdf results,}
\hl{and \app{appsect:ablations} contains additional ablation experiments.}}

\vspace{-0.25em}
\subsection{Intra-Category Generalization}
\label{sec:intra-category}

To evaluate whether our method can generalize to shapes in the same object category, 
we train an \ngdf model on 7 shapes in the ``Bottle'' category from ACRONYM~\cite{acronym2020}.
Training samples are generated from the meshes using the same data collection procedure described in \sect{sec:impl-details}. 
We evaluate performance on a held-out Bottle instance, the same instance used in the previous evaluations. 
The intra-category model achieves $0.63$ execution success on 30 Bottle trials for the reaching and grasping evaluation, which is comparable with the single-object \ngdf results from Table~\ref{table:eval-reach-grasp}, demonstrating intra-category generalization without loss of performance.

\begin{figure}[ht]
    \centering
    \includegraphics[width=\columnwidth]{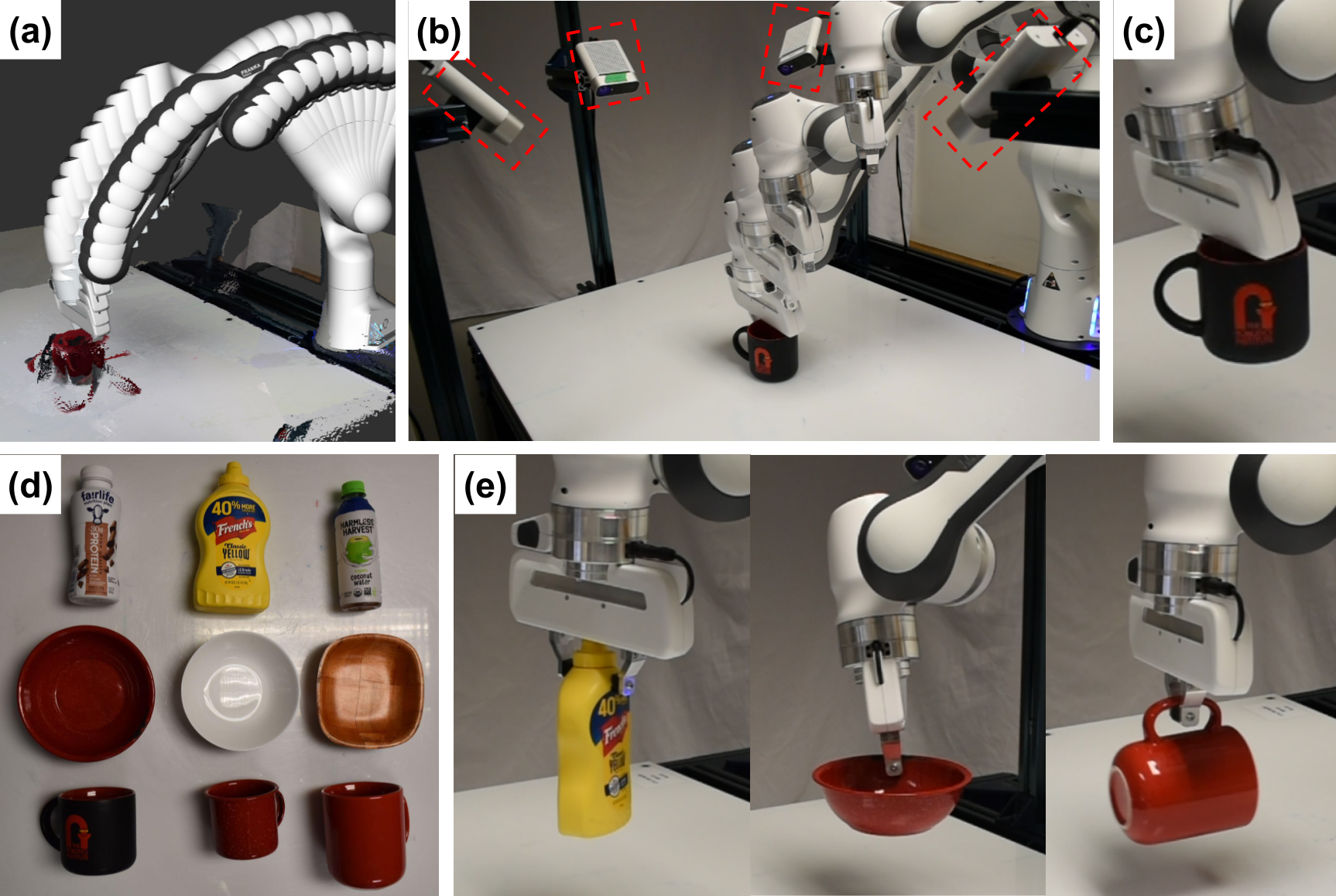}
    \caption{\hl{Real System Evaluation. (a) Visualizing the plan and imperfect object point cloud; (b) executing the plan on hardware (cameras highlighted with red boxes); (c) lifting the object. (d) The nine objects used for testing. (e) Additional successful grasps.}
    }
    \label{fig:real}
    \vspace{-1.5em}
\end{figure}




\vspace{-0.25em}
\subsection{\hl{Real Robot Reaching and Grasping Evaluation}}
\label{sec:real}

\hl{
Finally, we test our method's reaching and grasping performance on a real robot system. 
At the start of each trial, an object is placed in a random stable pose.
A partial point cloud of the scene is obtained from four Azure Kinect depth sensors (\fig{fig:real}b), similar to NDF~\cite{simeonovdu2021ndf}.
The object point cloud is segmented via plane fitting, then passed
as input to \ngdf models from~\sect{sec:eval-ngf}, which are trained on one instance per category in simulation.
The cloud is also converted to a signed distance field to enable computing collision costs with CHOMP~\cite{zucker2013chomp}.
The optimized trajectory is executed on a Franka Panda robot with impedance control, and the trial is considered successful if the object is grasped and lifted without being dropped (\fig{fig:real}c).
9 test objects were evaluated, 3 from each shape category (\fig{fig:real}d).
3 grasp attempts were performed per object for a total of 27 trials.
\appref{See~\app{appsect:real_impl}~\arxivref{in~\cite{weng2022neural}} for additional details.}
}

\hl{
Our overall grasp success rate was 81\%, with success per category being 7/9 Bottles, 9/9 Bowls, and 6/9 Mugs.
Our system successfully grasped every object, despite many of them being outside of its training distribution in terms of size and shape. 
The method also demonstrated robustness to noisy perception and 
execution with impedance control.
Failure cases were due to slight collisions between the fingers and objects, similar to what we observed in simulation.
}

\vspace{-0.3em}
\section{Discussion}
\label{sec:discuss}
\vspace{-0.1em}

Neural implicit functions have been widely explored for 3D vision tasks such as shape reconstruction.
\ngdf extends this concept to grasp estimation, using 6D poses as queries on grasp manifolds. 
Our work differs from existing work on 3D reconstruction, not only due to the higher dimensionality of our problem, but also because of the challenge in acquiring ground truth labels.
The ground truth grasp distance between an arbitrary query pose and the corresponding closest grasp is expensive to compute. 
Instead, we train on large-scale discrete grasp sets~\cite{acronym2020} as near-ground truth supervision. 
Our experiments in~\sect{sec:eval-ngf} show that \ngdf is able to learn the continuous grasp manifold as the level set of the neural field from this discrete supervision. 

\ngdf decouples the problem of learning a grasp manifold representation from the problem of finding a good grasp pose. 
For the latter, we formulate the distance output of \ngdf as a cost to be minimized. 
For the full robot motion planning regime, we jointly optimize the grasp cost with smoothness and collision costs.
We outperform baselines in~\sect{sec:eval-gmp} that represent what a practitioner would implement for a reaching and grasping task.
While the performance of oracle methods indicate room for improvement, our results show that joint optimization with \ngdf is a promising direction for manipulation. 
We also demonstrate scalability with intra-category generalization results in~\sect{sec:intra-category}, \hl{and deploy our method on real hardware in~\sect{sec:real}.}

In terms of limitations, 
\ngdf is trained on a gripper-specific dataset; \ngdf for other grippers \hl{may} require different datasets.
\hl{The method also depends on upstream object segmentation.}
Further, the cost weights are fixed during optimization in the reach and grasp planning task; learning to adjust the weights each iteration could improve performance.

\vspace{-0.3em}
\section{Conclusion}
\vspace{-0.1em}

We propose Neural Grasp Distance Fields (\ngdf), which represent the continuous manifold of grasps as the zero-level set of a neural field.
We formulate the estimated distance as a cost for a gradient-based trajectory optimizer to jointly optimize with other trajectory costs such as smoothness and collision avoidance to perform reach and grasp planning. 
Our results show that \ngdf outperforms existing methods, while generalizing to unseen poses and unseen objects. 

\vspace{-0.3em}
\section*{Acknowledgment}
\vspace{-0.1em}

\hl{This work was supported by the US Air Force and DARPA (FA8750-18-C-0092), NSF (IIS-1849154, DGE2140739), CMU GSA/Provost Conference Funding, and the Meta AI Mentorship Program. 
The authors thank Kalyan Alwala and Adithya Murali for early prototyping, as well as Taosha Fan, Austin Wang, Daniel Seita, and Chuer Pan for helpful discussions and feedback.
}

\appref{
\appendix
\beginsupplement

\subsection{Ablations for Neural Grasp Distance Fields}
\label{appsect:ablations}

We perform ablation experiments for our trajectory optimizer (Table~\ref{tab:ablation}).
We compare using Adam~\cite{Kingma2015AdamAM} vs. a fixed step size (``No-Adam'') for functional gradient descent.  
Unlike our method, CHOMP~\cite{zucker2013chomp} originally uses a fixed or decaying step size, in the setting where the start and end trajectory configurations are not optimized (\sect{sec:jgmp}). 
In our setting, the end configuration is variable to allow optimization of the grasp pose.
No-Adam converges slowly when the trajectory is far from a valid grasp pose, and overshoots when near the level set. 
We also evaluated using a decaying step size; while this mitigated the overshooting issue, convergence was still much slower, and the decay rate required tuning.

``No-Initial-IK'' initializes the configuration at every timestep in the trajectory to the starting joint configuration, instead of using IK to initialize the trajectory as described in~\sect{sec:eval-gmp}.
We observe worse performance with No-Initial-IK as the initial trajectory is farther from the desired grasp trajectory, making it harder to plan. 

\begin{table}[h]
    \centering
    \caption{\normalsize Optimizer Ablation Results}
    \label{tab:ablation}
    \small
    \begin{tabular}{l c}
        \toprule
        & Grasp Execution $\uparrow$ \\
        \midrule
        \ngdf, No-Adam
            & $0.18$ \\
        \ngdf, No-Initial-IK
            & $0.44$ \\
        \ngdf (Ours)
            & $\mathbf{0.61}$ \\
        \bottomrule
    \end{tabular}
    \vspace{0.5em}
    \caption*{
    No-Adam uses CHOMP~\cite{zucker2013chomp} with a fixed step size instead of Adam~\cite{Kingma2015AdamAM} optimization for the functional gradient update. 
    No-Initial-IK initializes the trajectory so all steps in the plan start at the initial joint configuration. 
    \ngdf uses Adam and initializes the endpoint of the trajectory using inverse kinematics to achieve the best performance. 90 trials were performed as in Table~\ref{table:eval-reach-grasp}.
    }
    \vspace{-2em}
\end{table}

\subsection{\hl{Simulation Experiment Details}}


\subsubsection{Camera Position in Simulation}

\fig{fig:sim_setup} shows the position of the four cameras in simulation.

\begin{figure}[ht]
    \centering
    \includegraphics[width=0.8\columnwidth]{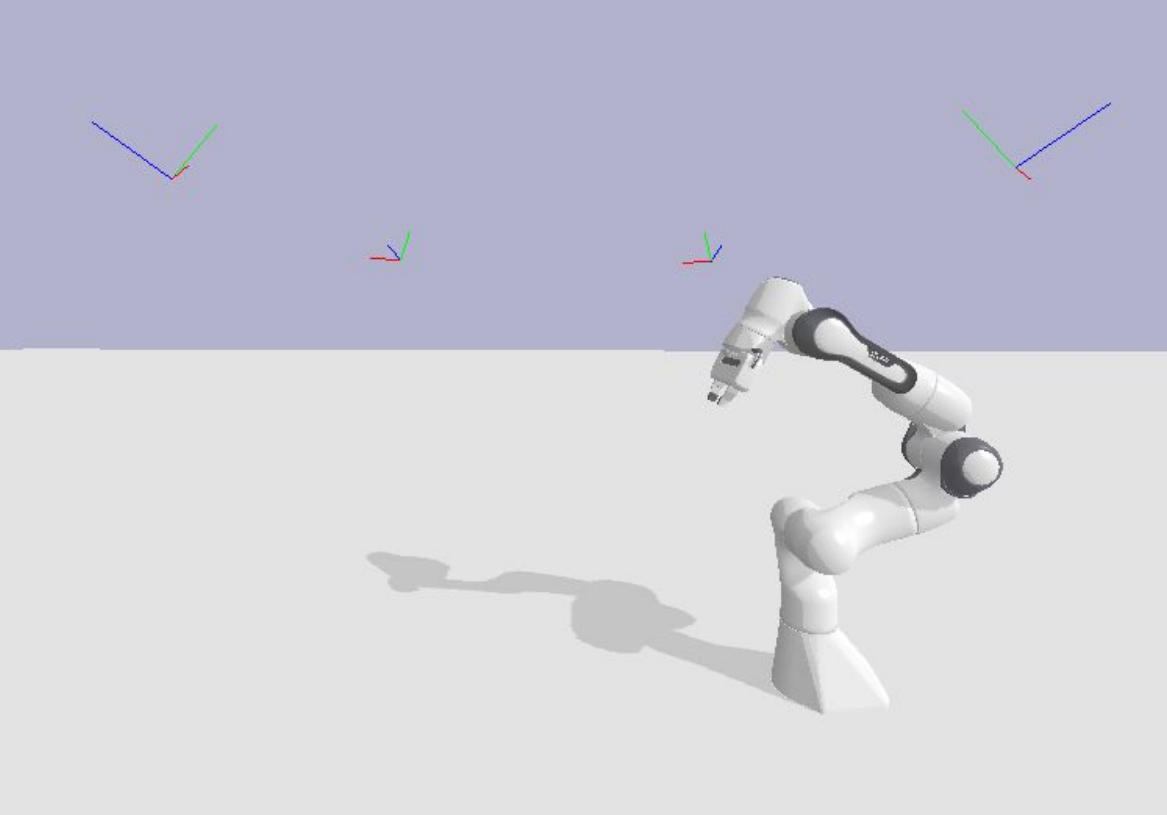}
    \caption{Camera poses in simulation visualized as axes. The negative z axis (in blue) is the camera optical axis and points toward the robot workspace.}
    \label{fig:sim_setup}
\end{figure}

\subsubsection{Qualitative Results}
\fig{fig:sim_traj} visualizes successful grasp trajectories in simulation for the reaching and grasping task from~\sect{sec:eval-ngf}. 

\begin{figure}[ht]
    \centering
    \begin{subfigure}[t]{\figw\textwidth}
        \includegraphics[width=\textwidth, trim={6cm 6cm 7cm 2cm},clip]{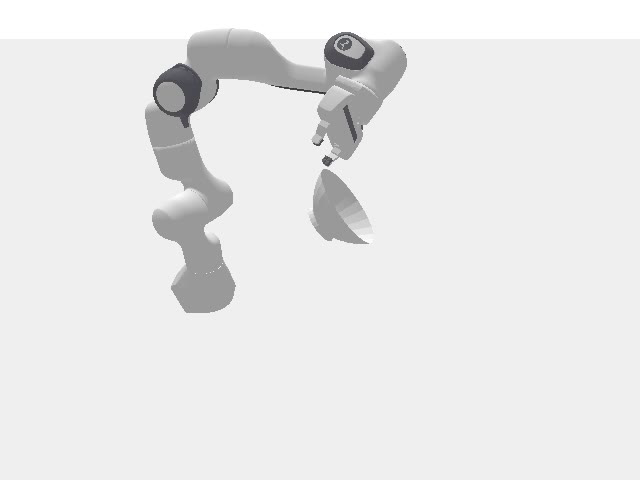}
    \end{subfigure}
    \begin{subfigure}[t]{\figw\textwidth}
        \includegraphics[width=\textwidth, trim={6cm 6cm 7cm 2cm},clip]{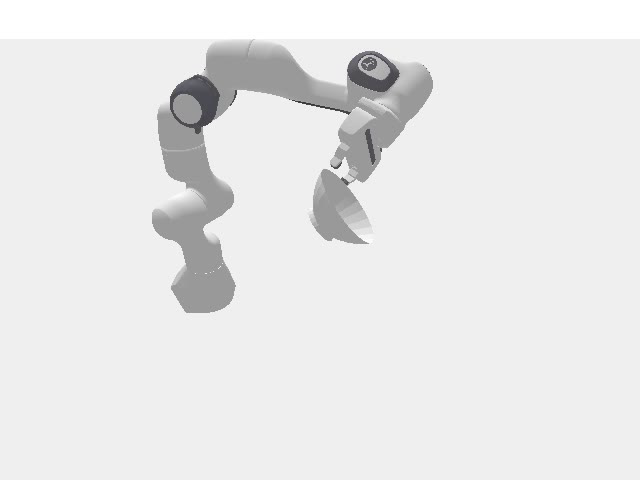}
    \end{subfigure}     \begin{subfigure}[t]{\figw\textwidth}
        \includegraphics[width=\textwidth, trim={6cm 6cm 7cm 2cm},clip]{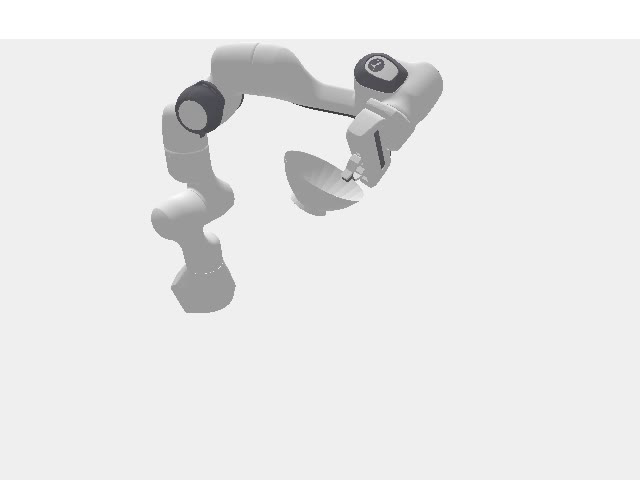}
    \end{subfigure} 
        \begin{subfigure}[t]{\figw\textwidth}
        \includegraphics[width=\textwidth, trim={6cm 6cm 7cm 2cm},clip]{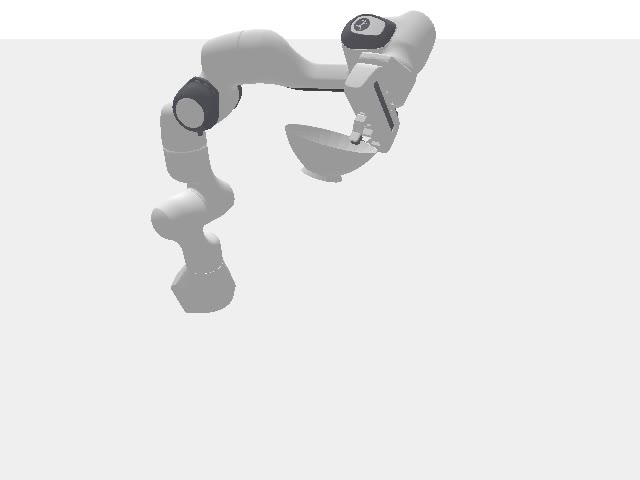}
    \end{subfigure} 
    \\
    \vspace{0.3em}
    \begin{subfigure}[t]{\figw\textwidth}
        \includegraphics[width=\textwidth, trim={5cm 6cm 9cm 2cm},clip]{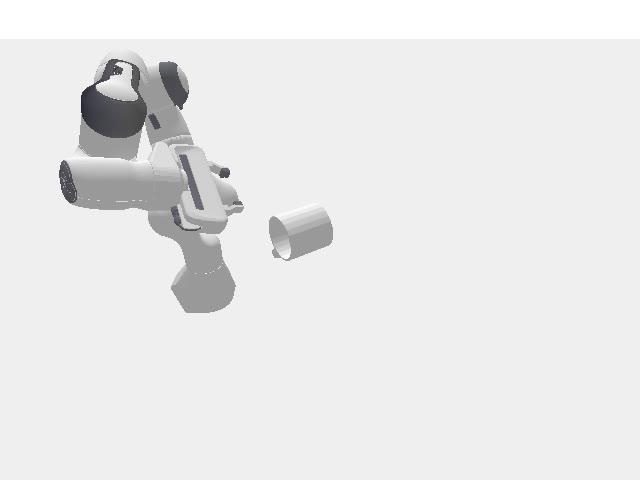}
    \end{subfigure}
    \begin{subfigure}[t]{\figw\textwidth}
        \includegraphics[width=\textwidth, trim={4cm 5cm 9cm 2cm},clip]{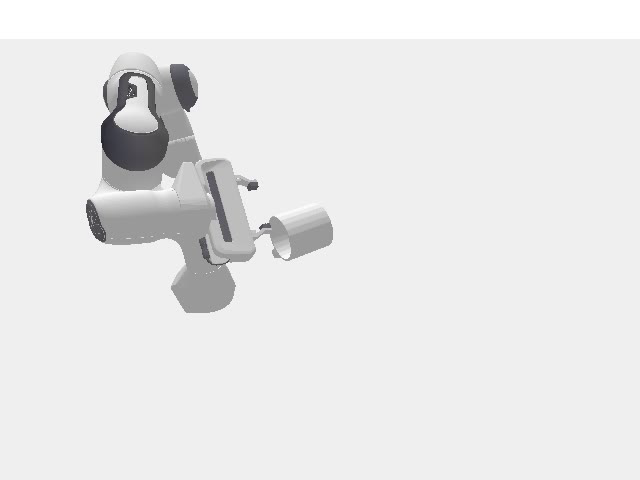}
    \end{subfigure}  
    \begin{subfigure}[t]{\figw\textwidth}
        \includegraphics[width=\textwidth, trim={4cm 5cm 9cm 2cm},clip]{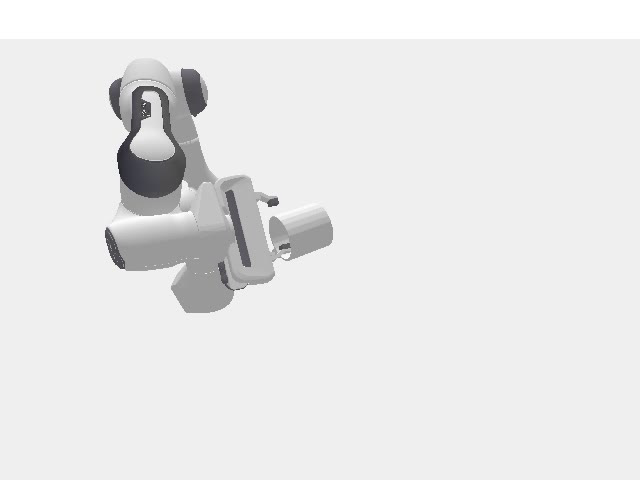}
    \end{subfigure} 
        \begin{subfigure}[t]{\figw\textwidth}
        \includegraphics[width=\textwidth, trim={4cm 5cm 9cm 2cm},clip]{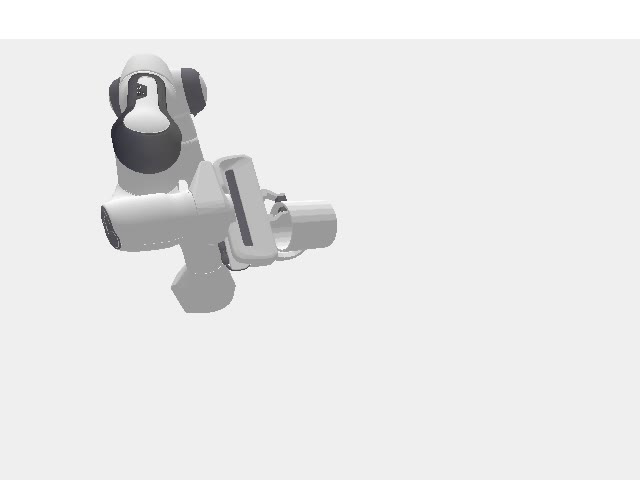}
    \end{subfigure} 
    \caption{Successful grasp trajectories (left-to-right) planned by our method for the bowl (top) and mug (bottom).}
    \label{fig:sim_traj}
\end{figure}

\subsection{\hl{Real System Experiment Details}}
\label{appsect:real_impl}

This section provides system implementation details for our real world experiments in~\sect{sec:real}. 
Our system consists of a 7-DOF Franka Panda robot and four Azure Kinect cameras (\fig{fig:real}b), a similar setup to NDF~\cite{simeonovdu2021ndf}.

\subsubsection{Calibration}

The Azure Kinect cameras were extrinsically calibrated using ColoredICP~\cite{park2017colored}.
For camera intrinsics, the factory calibration was used. 
Robot-camera extrinsic calibration was performed using Tsai-Lenz~\cite{tsai1989new}.
The calibrated cameras produce a combined scene point cloud in the robot base frame. 

\subsubsection{Point Cloud Processing}
\label{sec:pcl_processing}

To segment the object point cloud from the scene point cloud, we fit a table plane using RANSAC and remove points belonging to the plane.
Outlier removal and DBScan~\cite{ester1996density} are used to refine the object point cloud. 
Our planner requires a signed distance field (SDF) of the object for collision avoidance, so we construct a mesh from the object point cloud, then compute the SDF from the mesh using the tools provided in Wang~\etal\cite{wang2020manipulation}. 

Even with four cameras, careful calibration, and point cloud processing, we recover partial point clouds with  inaccuracies and noise (see ~\fig{fig:noise}).
Despite these deficiencies, our method achieves a high success rate on real objects in various configurations (\sect{sec:real}), demonstrating robustness to perceptual errors. 

\begin{figure*}[t]
    \centering
    \begin{subfigure}[t]{0.32\textwidth}
        \centering
        \includegraphics[width=0.8\textwidth]{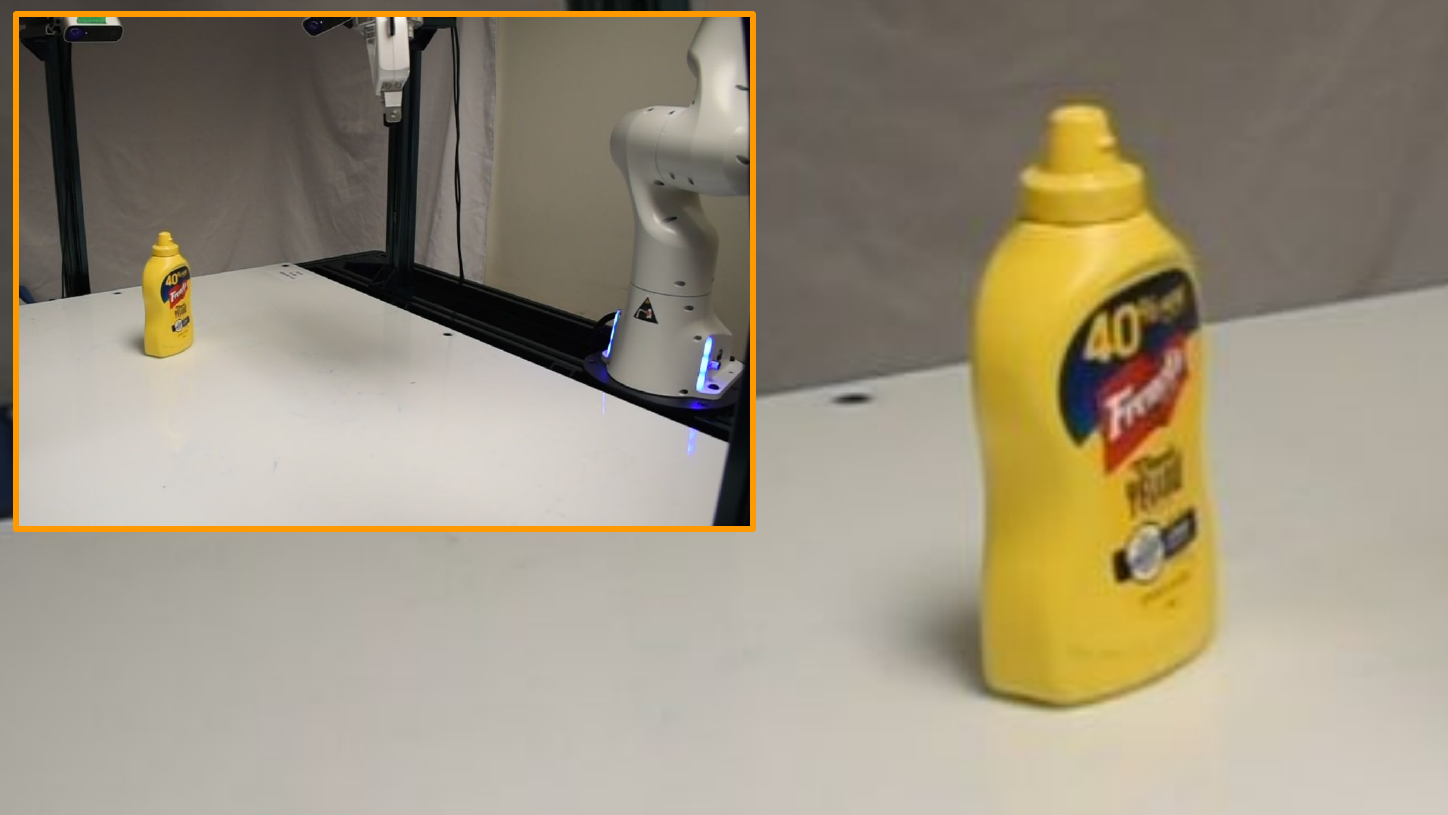}
        \caption{Bottle}
    \end{subfigure}
    \begin{subfigure}[t]{0.32\textwidth}
        \centering
        \includegraphics[width=0.8\textwidth, trim={17cm 9cm 17cm 7cm},clip]{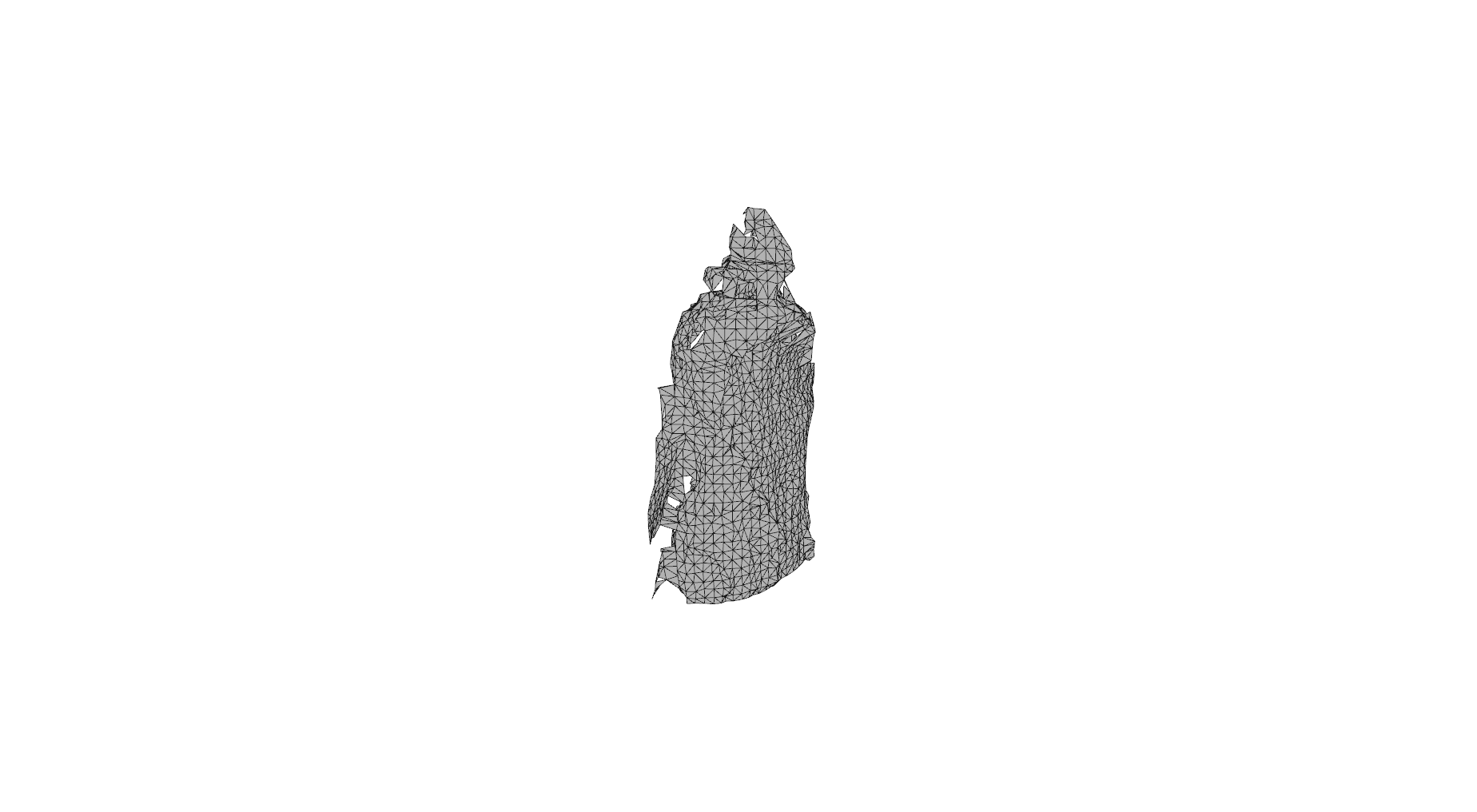}
        \caption{Reconstructed Bottle Mesh, View 1}
    \end{subfigure}
        \begin{subfigure}[t]{0.32\textwidth}
        \centering
        \includegraphics[width=0.8\textwidth, trim={10cm 4cm 10cm 4cm},clip]{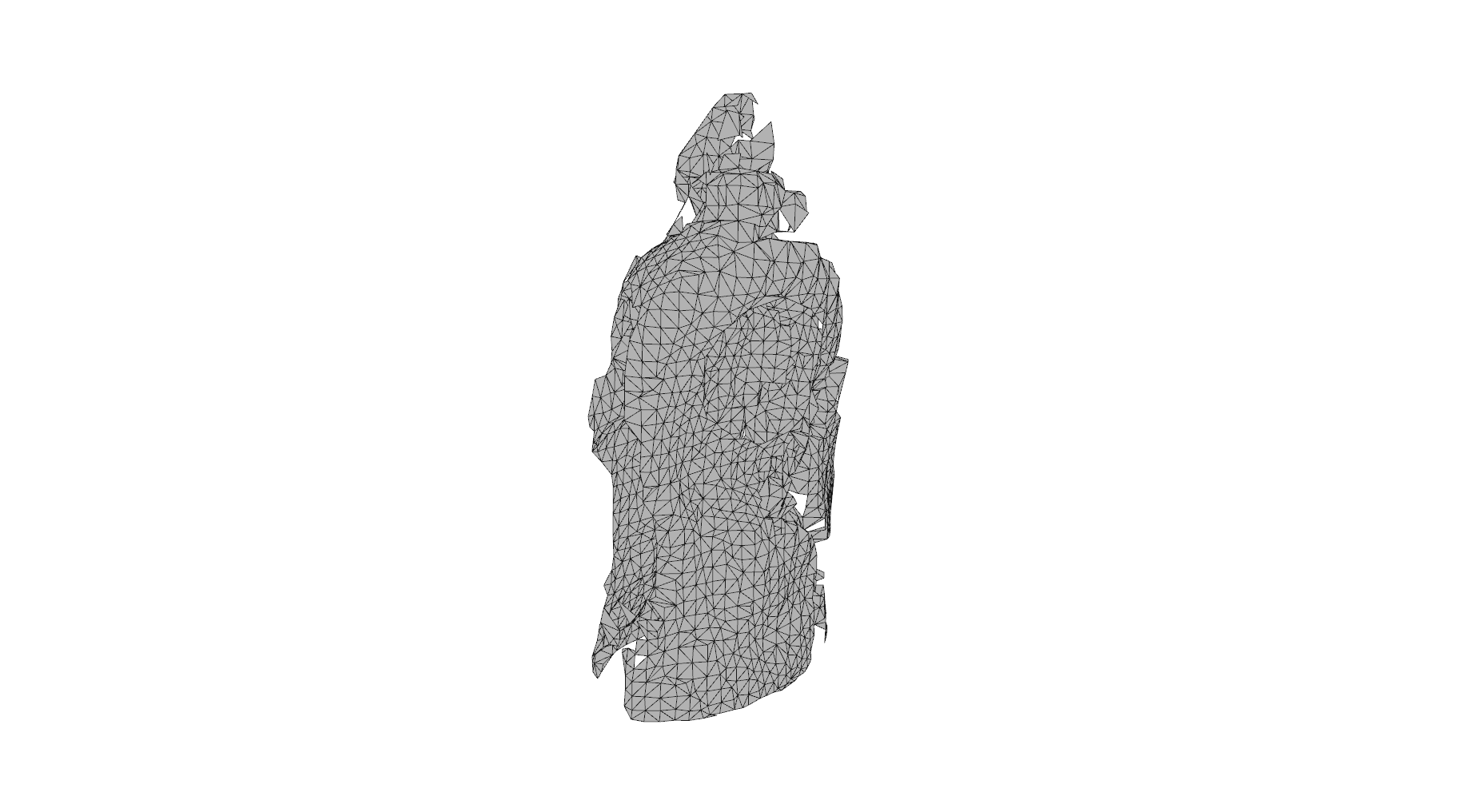}
        \caption{Reconstructed Bottle Mesh, View 2}
    \end{subfigure}
    \begin{subfigure}[t]{0.32\textwidth}
        \centering
        \includegraphics[width=0.8\textwidth, ]{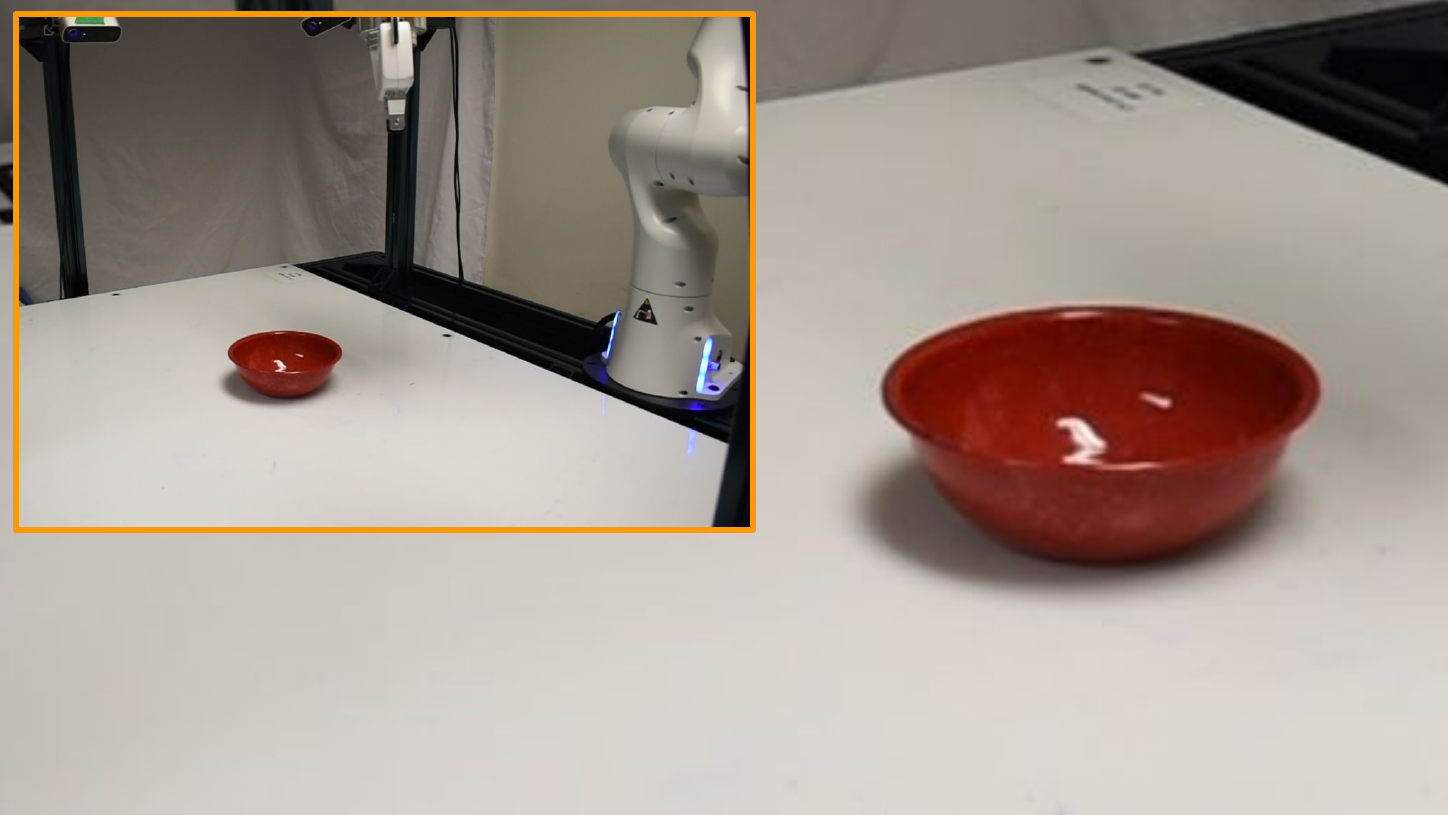}
        \caption{Bowl}
    \end{subfigure}
    \begin{subfigure}[t]{0.32\textwidth}
        \centering
        \includegraphics[width=0.8\textwidth, trim={15cm 7cm 15cm 7cm},clip]{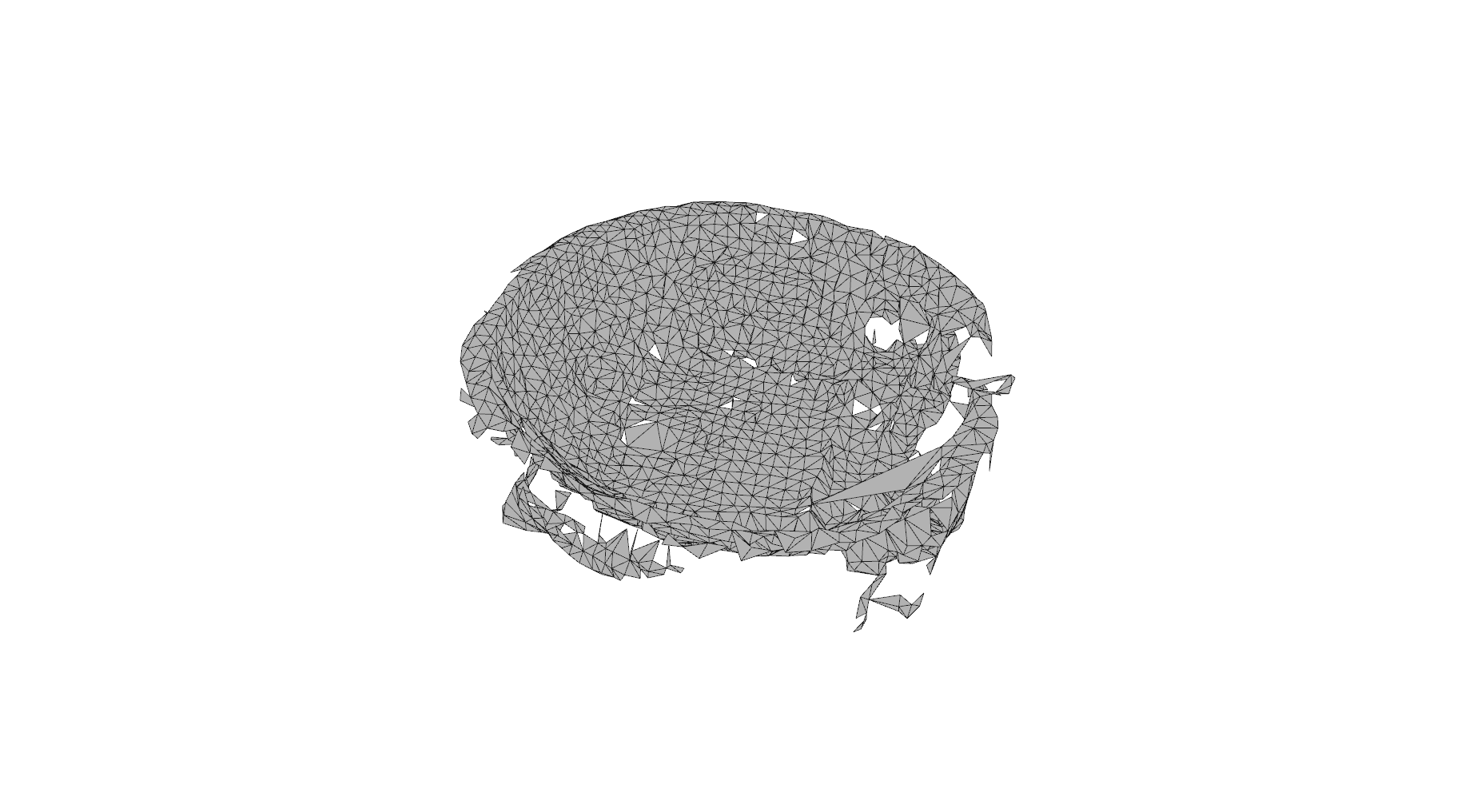}
        \caption{Reconstructed Bowl Mesh, View 1}
    \end{subfigure}
    \begin{subfigure}[t]{0.32\textwidth}
        \centering
        \includegraphics[width=0.8\textwidth, trim={12cm 5cm 12cm 5cm},clip]{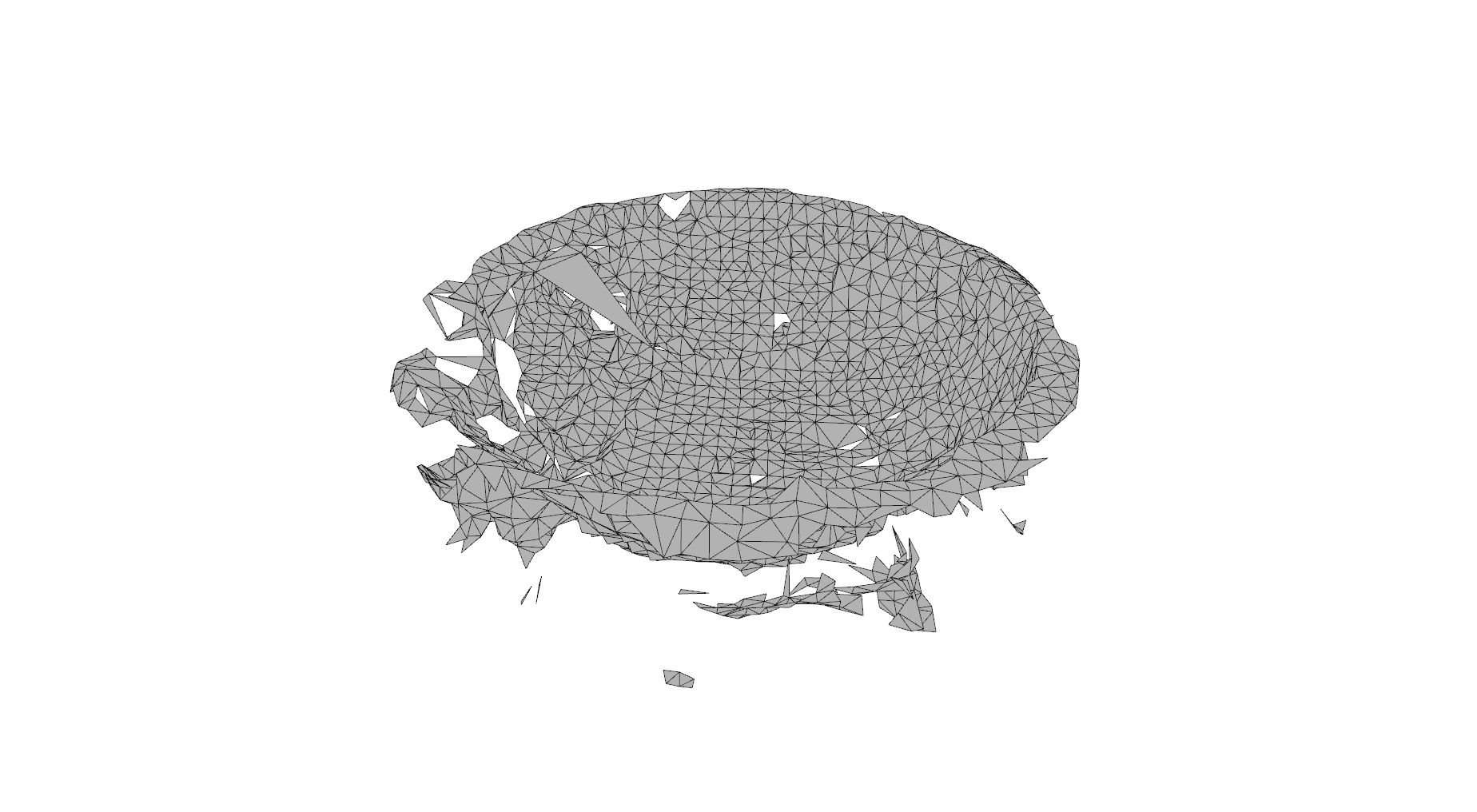}
        \caption{Reconstructed Bowl Mesh, View 2}
    \end{subfigure}
    \begin{subfigure}[t]{0.32\textwidth}
        \centering
        \includegraphics[width=0.8\textwidth]{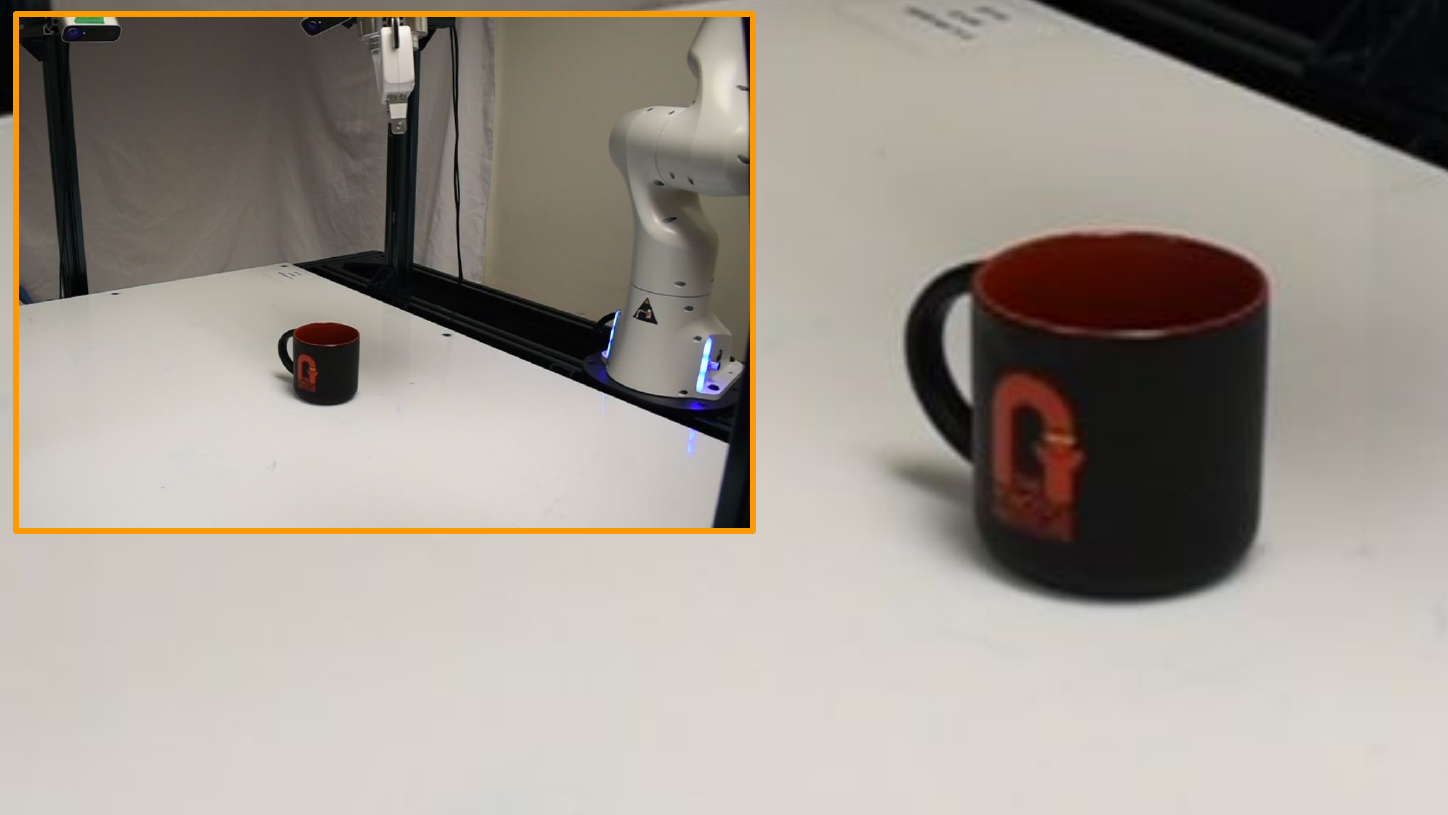}
        \caption{Mug}
    \end{subfigure}
    \begin{subfigure}[t]{0.32\textwidth}
        \centering
        \includegraphics[width=0.8\textwidth, trim={15cm 7cm 15cm 7cm},clip]{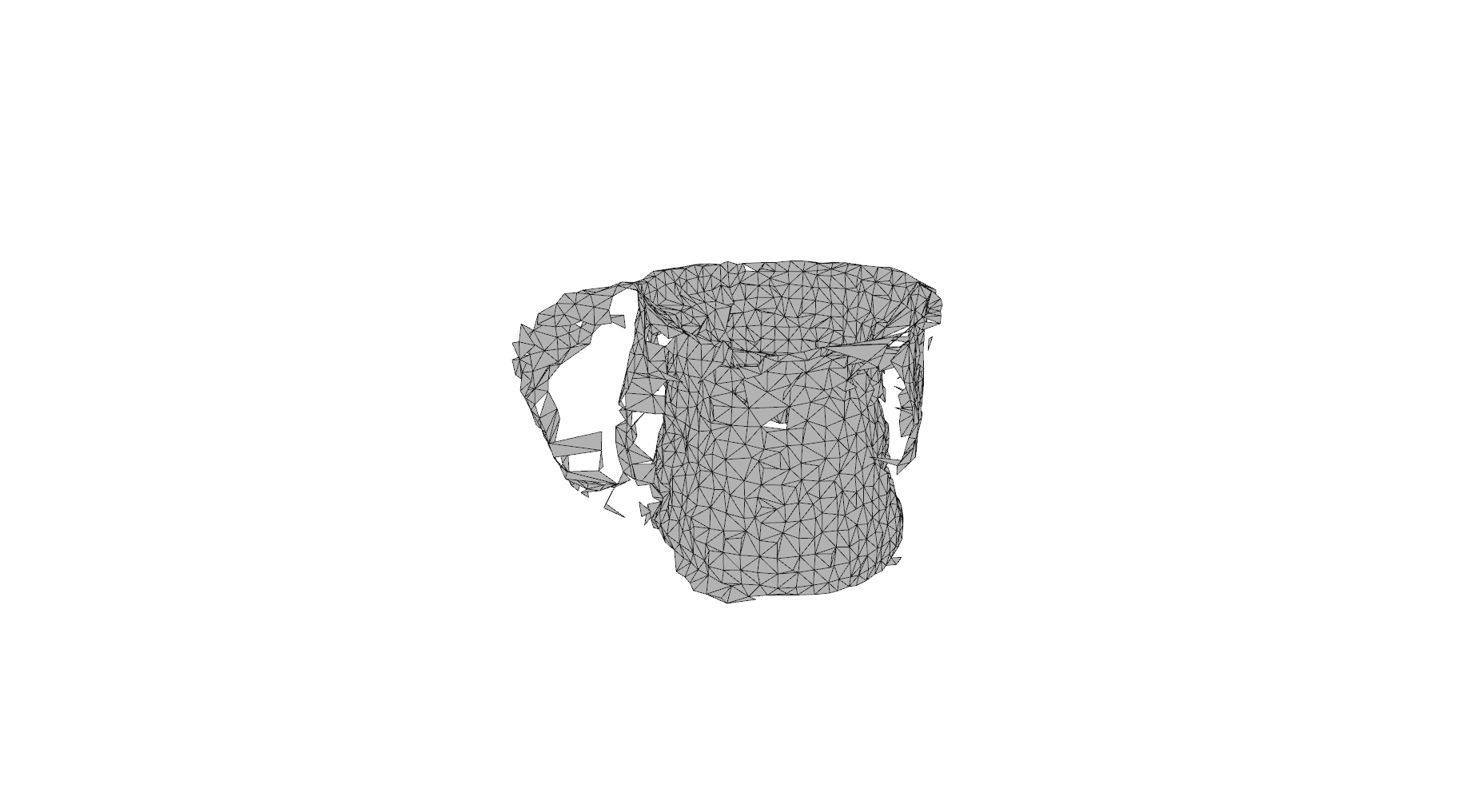}
        \caption{Reconstructed Mug Mesh, View 1}
    \end{subfigure}
    \begin{subfigure}[t]{0.32\textwidth}
        \centering
        \includegraphics[width=0.8\textwidth, trim={8cm 5cm 8cm 5cm},clip]{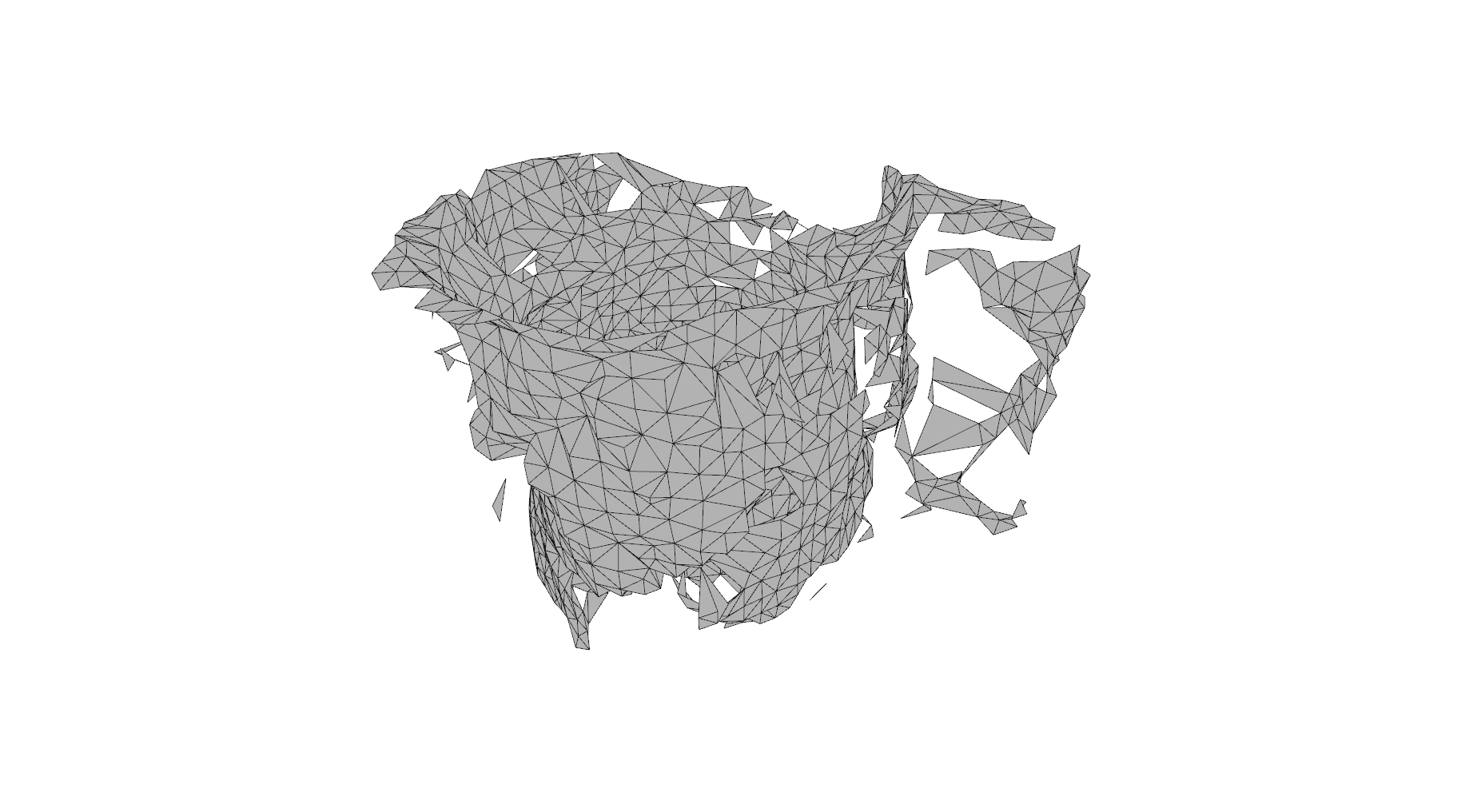}
        \caption{Reconstructed Mug Mesh, View 2}
    \end{subfigure}
    \caption{
    Meshes reconstructed during system evaluations.
    The first column shows the placement of the objects in each trial, along with a close-up image of the object itself. 
    The second column shows the meshes reconstructed from the four depth cameras according to the procedure in 
    \sect{sec:pcl_processing}, posed roughly as they appear in the first column.
    The third column shows the back of each mesh.
    Note that these meshes and images are magnified for visual clarity and are not consistently scaled.
    Even with outlier removal and other mesh processing techniques, we observe inaccuracies in the reconstruction; however, our method is robust to these inaccuracies as demonstrated by our results in \sect{sec:real}.}
    \label{fig:noise}
\end{figure*}

\subsubsection{Control}
The output of the planner is a joint angle trajectory consisting of 30 timesteps. In order to execute the trajectory on the Franka Panda, the total duration of trajectory execution is set to 5 seconds, and the trajectory is interpolated using cubic spline interpolation to provide joint angles at \SI{1}{\hertz}. Impedance control~\cite{zhang2020modular} is then used to execute the high-frequency trajectory.

\subsection{Experimental Setup}

\begin{table}[h]
    \centering
    \caption{\normalsize Real Object Pose Configurations}
    \label{tab:configs}
    \small
    \begin{tabular}{l c}
        \toprule
        & Sampled Pose Configurations \\
        \midrule
        Bottle 1 (Protein Drink)
            & Upright, Sideways \\
        Bottle 2 (Mustard Bottle)
            &  Upright \\
        Bottle 3 (Coconut Water)
            & Upright, Sideways \\
        Bowl 1 (YCB Bowl)
            & Upright \\
        Bowl 2 (White Bowl)
            & Upright \\
        Bowl 3 (Square Bowl)
            & Upright \\
        Mug 1 (Black Mug)
            & Upright, Sideways, Upside Down \\
        Mug 2 (YCB Mug) 
            & Upright, Sideways, Upside Down \\
        Mug 3 (Large Mug) 
            & Upright, Sideways, Upside Down \\
        \bottomrule
    \end{tabular}
    \vspace{0.5em}
    \caption*{The sampled pose configurations for objects in the real system evaluation. Objects are numbered left to right according to \fig{fig:real}d. Some stable poses did not permit grasping and were omitted; for example, the mustard bottle cannot be grasped lying sideways as it is too wide.
    }
    \vspace{-1em}
\end{table}

Several of the test objects could be grasped via multiple stable pose configurations. 
For example, a mug can be grasped while upright, on its side, or upside down.
For objects with multiple graspable pose configurations, the configuration was randomly sampled for each trial. The 9 test objects (see \fig{fig:real}d) had graspable stable pose configurations shown in Table~\ref{tab:configs}.}

\balance
\bibliographystyle{IEEEtran.bst}
\bibliography{references}

\end{document}